\documentclass[10pt,twocolumn,letterpaper]{article}
\usepackage[pagenumbers]{cvpr}
\usepackage{amsmath,amssymb,booktabs,array}
\usepackage{colortbl}
\definecolor{cvprblue}{rgb}{0.21,0.49,0.74}
\definecolor{paperred}{HTML}{BF4D55}
\definecolor{paperredlight}{HTML}{FBEDEF}
\usepackage[breaklinks,colorlinks,allcolors=cvprblue]{hyperref}
\def\confName{CVPR}
\def\confYear{2027}
\newcommand{\chroma}{\mathbf{c}}
\newsavebox{\eventseecolortitlebox}
\newcommand{\eventseecolortitle}{%
  \sbox{\eventseecolortitlebox}{EventSeeColor}%
  \raisebox{-\dp\eventseecolortitlebox}{%
    \includegraphics[width=\wd\eventseecolortitlebox,
      height=\dimexpr\ht\eventseecolortitlebox+\dp\eventseecolortitlebox\relax]{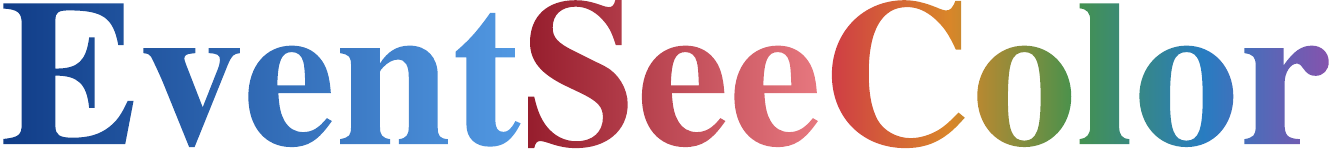}}}
\title{\texorpdfstring{\eventseecolortitle}{EventSeeColor}: Learning to Reconstruct Color from Monochrome Events}
\hypersetup{pdftitle={EventSeeColor: Learning to Reconstruct Color from Monochrome Events},
  pdfauthor={Chuanzhi Xu, Langyi Chen, Haoxian Zhou, Zelin Zhang, Haodong Chen, Qiang Qu, Xiaoming Chen, Weidong Cai},
  pdfsubject={arXiv preprint}}
\author{%
Chuanzhi Xu\textsuperscript{1,*}\quad Langyi Chen\textsuperscript{1}\quad Haoxian Zhou\textsuperscript{1}\quad Zelin Zhang\textsuperscript{1}\\
Haodong Chen\textsuperscript{1}\quad Qiang Qu\textsuperscript{1}\quad Xiaoming Chen\textsuperscript{2}\quad Weidong Cai\textsuperscript{1}\\[3pt]
{\normalsize\textsuperscript{1}The University of Sydney\quad\textsuperscript{2}Beijing Technology and Business University\quad\href{https://github.com/LouckXu/EventSeeColor}{\raisebox{-0.12em}{\includegraphics[height=0.9em]{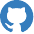}}\hspace{0.3em}Project Page}}}

\begin{document}
\raggedbottom
\twocolumn[{
  \renewcommand\twocolumn[1][]{#1}
  \maketitle
  \begin{center}
    \captionsetup{hypcap=false}
    \includegraphics[width=\textwidth]{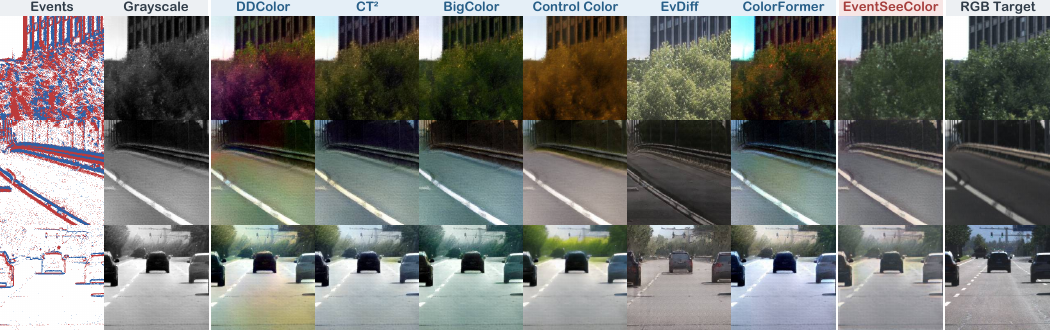}
    \captionof{figure}{Color reconstruction from monochrome events showing foliage, road markings, and vehicles. EventSeeColor uses reconstructed grayscale and aligned events to recover scene appearance with more faithful colors and finer details than the baselines.}
    \label{fig:teaser}
  \end{center}
  \vspace{0.4em}
}]

\begingroup
\renewcommand{\thefootnote}{*}
\footnotetext{Corresponding author: \href{mailto:chuanzhi.xu@sydney.edu.au}{\texttt{chuanzhi.xu@sydney.edu.au}}.}
\endgroup

\renewcommand{\endabstract}{\par}
\begin{abstract}
Event cameras capture visual brightness changes under challenging lighting conditions. However, traditional event-based image reconstruction targets grayscale intensity and produces images without scene color. Color distinguishes surfaces with similar intensity. Reconstructing color from events can reveal appearance differences that reconstructed grayscale images cannot express. Although monochrome event cameras do not directly measure color, their event-triggering patterns depend on surface reflectance, illumination, and the sensor's spectral response. These patterns may contain cues that support color reconstruction. In this paper, we present EventSeeColor, which uses event-reconstructed grayscale and aligned events to jointly guide a diffusion colorizer. Grayscale-derived semantic context supports color inference, while region-balanced chroma supervision and local RGB refinement improve color distribution, brightness, and texture. In the combined evaluation on two real event datasets, EventSeeColor achieves the best structural similarity, learned perceptual scores, and chromatic-region color difference among the external methods.
\end{abstract}

\section{Introduction}
\label{sec:intro}
Event cameras asynchronously record changes in light intensity, providing high temporal resolution and a wide dynamic range~\cite{Gallego2022Survey}, which support visual sensing during rapid motion and challenging lighting conditions. To make such observations accessible to people and frame-based vision models, traditional event-based image reconstruction converts sparse temporal measurements into grayscale intensity images~\cite{Rebecq2019,V2VE2VID2025}. These images recover scene structure but omit color differences that can be important for interpreting objects and surfaces~\cite{Tanaka1999ColorDiagnosticity}. For example, differently colored road markings or vehicle surfaces may have similar grayscale intensity. As shown in Figure~\ref{fig:teaser}, reconstructing their color can reveal appearance distinctions that grayscale reconstruction cannot express, enriching the scene information from monochrome event sensing.

The event formation process helps explain the connection between event activity and scene color. Although monochrome event cameras do not directly measure color, their measured intensity depends on incident light and the sensor's spectral response~\cite{Arja2025MonochromeSpectrum}. As motion and illumination change this intensity, its temporal variation determines event timing and polarity~\cite{Gallego2022Survey}. Therefore, event activity is connected to the image formation process that determines scene color. This relationship suggests that event-triggering patterns may contain useful cues for color inference. It motivates a central question: \textit{how can temporal event cues and reconstructed spatial structure jointly guide color reconstruction from ordinary monochrome recordings?}

Image colorization models provide an intuitive starting point by inferring plausible colors from the semantic content of a grayscale image~\cite{Zhang2016,Kang2023}. Applying these models to an event-reconstructed grayscale image offers a straightforward approach to color reconstruction. However, colorizers that rely solely on semantic cues from event-reconstructed grayscale without access to temporal event information may struggle to reconstruct color faithfully, as shown in Figure~\ref{fig:teaser}.

Recent event-driven diffusion models also use generative priors to reconstruct color videos from event observations~\cite{Liang2024E2VIDiff,Chen2024LaSeE2V,EvDiff2025}. These methods reconstruct complete video frames by jointly generating scene structure and color. Sparse events do not fully determine scene appearance, and generative priors can introduce hallucinated details and reconstruction artifacts~\cite{Liang2024E2VIDiff,Chen2024LaSeE2V}. Moreover, existing methods do not explicitly examine whether aligned events improve color accuracy beyond the spatial and semantic guidance of reconstructed grayscale images.

In this paper, we present EventSeeColor, to our knowledge the first framework for event-guided colorization of event-reconstructed grayscale images, with three core components. \textit{Event and Grayscale Fusion} combines spatial and semantic cues from reconstructed grayscale with temporal cues from aligned events to guide a diffusion colorizer. \textit{Balanced Chroma Refinement} supports regional color recovery, and \textit{Local Detail Refinement} corrects brightness, color, and texture in the rendered RGB image. All scene-dependent inference inputs originate from the monochrome event stream. We evaluate EventSeeColor against image colorizers and direct event-to-video methods on paired real event recordings from PEOD~\cite{Cui2026} and CoSEC~\cite{Peng2024}. In the combined evaluation, EventSeeColor achieves the best structural similarity, perceptual fidelity, and chromatic-region color accuracy. Ablation studies show how aligned events, the color prior, and local RGB refinement contribute to color reconstruction. Our contributions are summarized as follows:
\begin{itemize}
  \item We propose EventSeeColor, to our knowledge the first framework for event-guided colorization of event-reconstructed grayscale images, jointly using the spatial and semantic content of reconstructed grayscale images and temporal cues from aligned monochrome events.
  \item We develop a sequential refinement pipeline that combines regional chroma supervision with local RGB correction to improve color placement, preserve neutral appearance, and reduce reconstruction artifacts.
  \item Experiments show EventSeeColor's strong structural, perceptual, and chromatic-region performance against external colorization and event reconstruction methods.
\end{itemize}

\begin{figure*}[t]
  \centering
  \includegraphics[width=\textwidth,pagebox=cropbox]{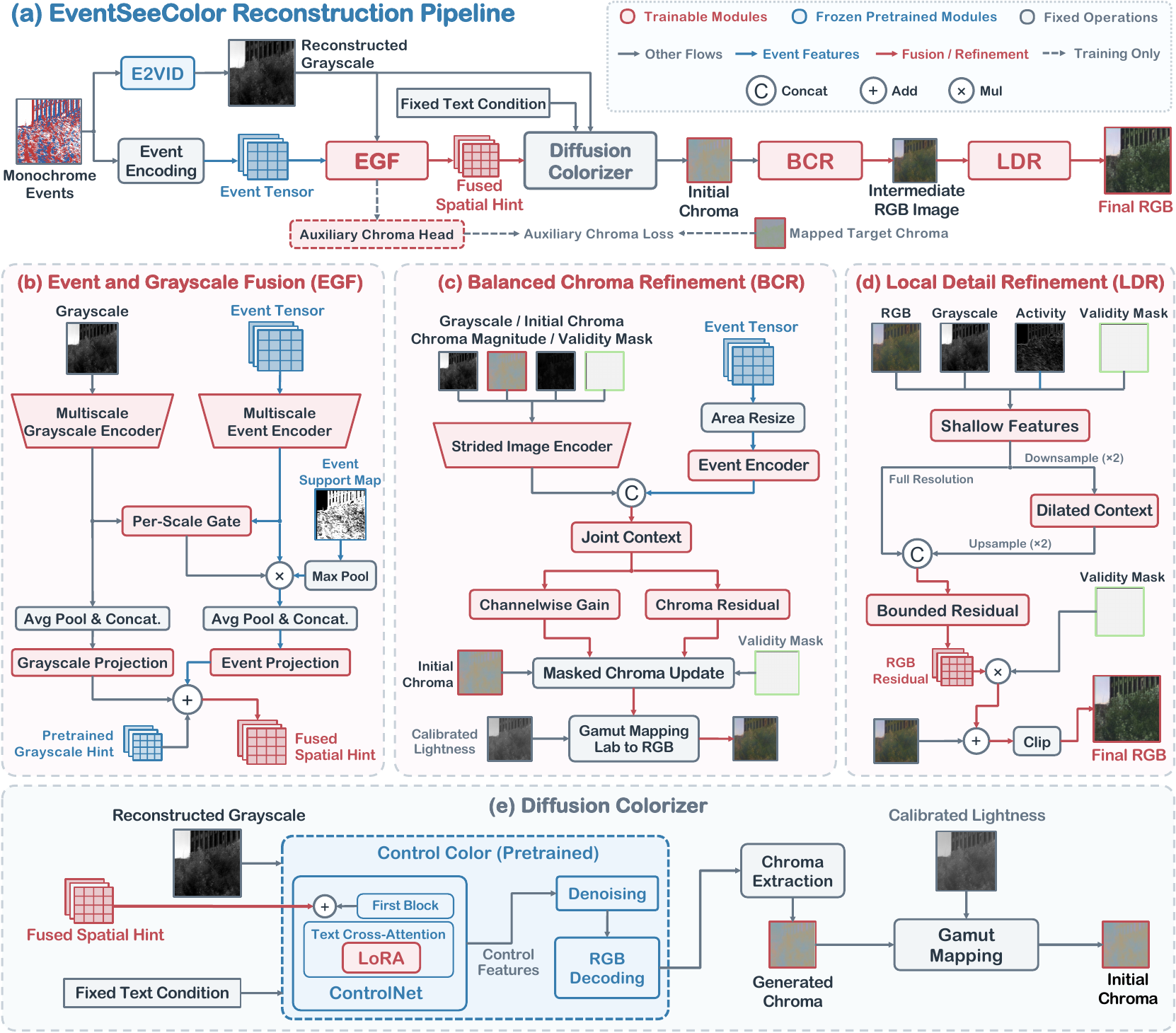}
  \caption{EventSeeColor architecture. (a) Overall pipeline. (b) EGF fuses grayscale and event features. (c) BCR refines chroma and renders intermediate RGB. (d) LDR predicts a bounded RGB residual. (e) The diffusion colorizer uses ControlNet for spatial guidance and low-rank adaptation (LoRA) to adapt text cross-attention. EGF, BCR, and LDR apply validity masks before encoding.}
  \label{fig:overview}
\end{figure*}

\section{Related Work}
\label{sec:related}
\noindent{\textbf{Grayscale Reconstruction.}}
Traditional event-based image reconstruction estimates grayscale intensity from monochrome events. E2VID~\cite{Rebecq2019} and V2V-E2VID~\cite{V2VE2VID2025} learn this mapping from event sequences, while Event-Diffusion~\cite{Liang2023EventDiffusion} and temporal residual diffusion~\cite{Zhu2024TemporalResidual} combine initial intensity reconstructions with event information for diffusion-based restoration. However, both groups focus on intensity recovery and do not address color differences between surfaces with similar grayscale intensity. EventSeeColor builds on the reconstructed spatial structure and uses aligned events to guide color inference, extending the recovered scene appearance beyond grayscale intensity.

\noindent{\textbf{Event-Based Color and Spectral Sensing.}}
Event-based color and spectral sensing introduces wavelength-dependent measurements through sensor, optical, or illumination design. Color-filtered events are used in RGBW sensor studies~\cite{Moeys2017ColorTemporal}, the CED dataset~\cite{Scheerlinck2019}, and Color4E~\cite{Ma2024Color4E}. Active RGB illumination~\cite{Cohen2023DVSColor}, E-RGB-D~\cite{Bajestani2026ERGBD}, and active hyperspectral imaging~\cite{Yu2025ActiveHyperspectral} combine event cameras with controlled illumination, while chromatic optics enable event-based spectral discrimination~\cite{Arja2025MonochromeSpectrum}. However, ordinary monochrome event recordings lack these additional spectral measurements. This is also one motivation for EventSeeColor: reconstructing color from monochrome events without additional spectral measurements at inference.

\noindent{\textbf{Grayscale Image Colorization.}}
Grayscale image colorization infers chroma from grayscale images and learned appearance priors. Colorful Image Colorization~\cite{Zhang2016}, DDColor~\cite{Kang2023}, ColorFormer~\cite{Ji2022ColorFormer}, and CT$^2$~\cite{CT22022} learn image representations for chroma prediction. BigColor~\cite{Kim2022BigColor} and DeOldify~\cite{DeOldify} use generative priors, while UniColor~\cite{UniColor2022} and DISCO~\cite{DISCO2022} incorporate appearance guidance. Palette~\cite{Saharia2022Palette}, L-CAD~\cite{LCAD2023}, and Control Color~\cite{Liang2025ControlColor} formulate colorization through diffusion. These methods can take event-reconstructed grayscale images as input and colorize them. However, their colorizers do not directly access event timing, polarity, or activity. Therefore, plausible colors inferred from image semantics may differ from the observed scene. EventSeeColor supplies aligned-event features as additional color guidance and uses regional chroma supervision to recover colored surfaces while preserving neutral appearance.

\noindent{\textbf{RGB Video Reconstruction from Events.}}
RGB video reconstruction includes methods with and without external color references. Pini et al.~\cite{Pini2020EventRGB}, ContinuityCam~\cite{Wang2025}, IE2Video~\cite{IE2Video2025}, and Engram-E2VID~\cite{Engram2026} reconstruct color videos from events and RGB reference images. E2VIDiff~\cite{Liang2024E2VIDiff}, LaSe-E2V~\cite{Chen2024LaSeE2V}, EvDiff~\cite{EvDiff2025}, and UniE2F~\cite{UniE2F2026} instead use diffusion priors to reconstruct RGB frames without an external color reference. LongE2V extends event-driven diffusion to long-horizon reconstruction, prediction, and interpolation~\cite{LongE2V2026}. However, generating scene structure and color from sparse events can introduce hallucinated details and artifacts~\cite{Liang2024E2VIDiff,Chen2024LaSeE2V}. These methods also do not directly evaluate whether events improve grayscale image colorization. EventSeeColor instead uses aligned events to guide grayscale image colorization and evaluates their impact on color accuracy.

\section{Methodology}
\label{sec:method}
As shown in Figure~\ref{fig:overview}, EventSeeColor reconstructs a color image $\widehat I_t$ from a monochrome event stream at time $t$. E2VID~\cite{Rebecq2019} provides a normalized grayscale image $G_t$, and the corresponding event interval provides an aligned tensor $V_t$. A validity mask $m_t$ identifies pixels with reliable alignment. Event and Grayscale Fusion (EGF) supplies a spatial condition to a diffusion colorizer. Balanced Chroma Refinement (BCR) adjusts the generated color and renders an intermediate RGB image $I_t^r$ using calibrated event lightness. Finally, Local Detail Refinement (LDR) predicts a local RGB correction to refine scene appearance. All scene-dependent inference inputs originate from the event stream. RGB images provide training supervision and evaluation targets. Appendix~\ref{app:inputs} specifies reconstruction, event encoding, and calibration.

\subsection{Event and Grayscale Fusion}
\label{sec:egf}
A grayscale reconstruction provides a spatial reference for color placement but does not explicitly retain event timing and polarity. EGF combines grayscale structure with the aligned event tensor at multiple scales so that temporal activity can guide color inference. Let $F_\ell^G$ and $F_\ell^E$ be grayscale and event features at scale $\ell$. The activity map $A_t$ averages the normalized positive and negative event counts, and max pooling its nonzero support gives $S_\ell$. A learned gate selects event features using both inputs:
\begin{equation}
  \widetilde F_\ell^E=
  \sigma\!\left(Q_\ell[F_\ell^G,F_\ell^E]\right)
  \odot F_\ell^E\odot S_\ell,
  \label{eq:gate}
\end{equation}
where $Q_\ell$ is a learned projection, $\sigma$ denotes sigmoid, brackets denote channel concatenation, and $\odot$ denotes elementwise multiplication. The support map places event conditioning at observed activity, while the gate controls its contribution using grayscale context.

The multiscale features are pooled to the resolution of the pretrained grayscale hint $H_t^0$ and projected into its feature space:
\begin{equation}
\begin{aligned}
  H_t={}&H_t^0+P_G[\mathcal P(F_1^G),\ldots,\mathcal P(F_K^G)]\\
        &+P_E[\mathcal P(\widetilde F_1^E),\ldots,\mathcal P(\widetilde F_K^E)],
\end{aligned}
\label{eq:fusion}
\end{equation}
where $\mathcal P$ is adaptive average pooling, $K$ is the number of scales, and $P_G$ and $P_E$ are learned projections. Both projections start at zero, preserving the pretrained hint at initialization. The event projection has no bias, so locations without pooled event support receive only grayscale features. The fused hint guides the Control Color latent diffusion backbone through ControlNet~\cite{Liang2025ControlColor,Rombach2022,Zhang2023}. Appendix~\ref{app:prompt} details the fusion architecture.

A fixed photographic prompt $\pi$, shared across all images, supplies the text condition during training and inference. Appendix~\ref{app:prompt} gives the fixed prompt settings.

\subsection{Balanced Chroma Refinement}
\label{sec:bcr}
Diffusion can produce plausible colors with incorrect regional strength or hue. Increasing saturation alone can also introduce false color into neutral roads, buildings, and white objects. BCR adjusts local hue and chroma to recover colored surfaces while preserving neutral appearance. Let $\chroma_t^0=(a_t^0,b_t^0)$ be the generated Lab chroma after gamut mapping at calibrated event lightness $L_t$. BCR receives $G_t$, $V_t$, $\chroma_t^0$, its magnitude, and $m_t$. Separate image and event encoders supply features to a joint context network. Two heads then predict a channelwise gain $\boldsymbol\alpha_t$ and a chroma residual $\mathbf r_t$:
\begin{equation}
  \chroma_t^r=\chroma_t^0+
  m_t\odot\left[(\boldsymbol\alpha_t-\mathbf1)\odot\chroma_t^0+\mathbf r_t\right].
  \label{eq:bcr}
\end{equation}
The independent gains adjust the strength of the $a$ and $b$ channels, and the residual corrects local hue and chroma. Bounded heads begin with an identity mapping through zero initialization. The intermediate image is $I_t^r=\mathcal T(L_t,\Pi_{L_t}(\chroma_t^r))$, where $\Pi_{L_t}$ scales out-of-gamut chroma into the RGB gamut and $\mathcal T$ converts Lab to RGB. This intermediate rendering anchors color to reconstructed intensity structure before local RGB refinement. Appendix~\ref{app:refinement} gives the parameterization.

\subsection{Local Detail Refinement}
\label{sec:ldr}
Correcting chroma alone cannot remove brightness errors or texture artifacts inherited from grayscale reconstruction. Therefore, LDR learns a local RGB correction using $I_t^r$, $G_t$, $A_t$, and $m_t$. Shallow features preserve fine structures, while a downsampled context branch provides a wider receptive field. Their concatenation produces a bounded residual and the final image:
\begin{equation}
\begin{aligned}
  \Delta I_t&=\delta\tanh f_\psi([I_t^r,G_t,A_t,m_t]\odot m_t),\\
  \widetilde I_t&=I_t^r+m_t\odot\Delta I_t,\\
  \widehat I_t&=\operatorname{clip}(\widetilde I_t,0,1),
\end{aligned}
\label{eq:output}
\end{equation}
where $f_\psi$ denotes the local network with parameters $\psi$ and $\delta$ bounds each RGB channel correction. Its final convolution is initialized to zero, so training starts from $I_t^r$. Direct RGB refinement lets the output jointly adjust local lightness, texture, and color. The final image is evaluated with its predicted RGB values. Appendix~\ref{app:ldr} details the network.

\subsection{Training Objectives}
\label{sec:objectives}
Training follows the inference stages. Diffusion adaptation updates EGF, a training-only auxiliary chroma head, and low-rank adaptation (LoRA) parameters in ControlNet text cross-attention~\cite{Hu2022LoRA}. The pretrained weights remain fixed. The diffusion colorizer objective combines valid-area denoising error $\mathcal L_{\rm diff}$ with Auxiliary Chroma Supervision $\mathcal L_{\rm aux}$:
\begin{equation}
  \mathcal L_{\rm gen}=\mathcal L_{\rm diff}+\lambda_{\rm aux}\mathcal L_{\rm aux}.
  \label{eq:genloss}
\end{equation}
The auxiliary head predicts coarse target chroma from $H_t$, connecting the fused representation to the color reconstruction task. Appendix~\ref{app:objectives} defines both terms.

Large neutral regions can dominate an image-wide chroma loss and weaken supervision for smaller colored surfaces. With the diffusion colorizer frozen, BCR uses a Region-Balanced Chroma Objective to give each target region direct supervision. Let $\ell_{ji}$ be the normalized chroma regression error at pixel $i$ of image $j$ and $w_{ji}$ its target-chroma weight. The valid pixel set $\Omega_j$ is partitioned into neutral, transitional, and chromatic regions. For the nonempty regions $\mathcal R_j$ and batch size $B$, the global and region terms are:
\begin{equation}
\begin{aligned}
  \mathcal L_{\rm global}&=
  \frac{\sum_j\sum_{i\in\Omega_j}w_{ji}\ell_{ji}}{\sum_j|\Omega_j|},\\
  \mathcal L_{\rm region}&=\frac1B\sum_j\frac1{|\mathcal R_j|}
  \sum_{r\in\mathcal R_j}\frac{\sum_{i\in\Omega_{jr}}w_{ji}\ell_{ji}}{|\Omega_{jr}|},
\end{aligned}
\label{eq:regionloss}
\end{equation}
where $\Omega_{jr}$ denotes region $r$ in image $j$. Equal region weighting prevents large regions from dominating smaller colored surfaces. Chroma shortfall, neutral-chroma supervision, and warm-region supervision complete the objective:
\begin{equation}
\begin{aligned}
\mathcal L_{\rm BCR}={}&(1-\beta)\mathcal L_{\rm global}+\beta\mathcal L_{\rm region}\\
 &+\lambda_s\mathcal L_{\rm short}+\lambda_n\mathcal L_{\rm neutral}
  +\lambda_w\mathcal L_{\rm warm}.
\end{aligned}
\label{eq:bcrloss}
\end{equation}
The regional terms preserve weak coloration while giving colored surfaces direct supervision. All region assignments use training targets and are absent from inference. Their definitions and reductions appear in Appendix~\ref{app:bcrloss}.

Finally, the diffusion colorizer and BCR are frozen while LDR learns from paired RGB targets. RGB reconstruction and gradient losses supervise local appearance and edges. Region-balanced chroma regression maintains color accuracy, the LPIPS loss~\cite{Zhang2018LPIPS} supervises perceptual similarity, and a clipping penalty discourages out-of-range predictions:
\begin{equation}
\begin{aligned}
\mathcal L_{\rm LDR}={}&\lambda_{\rm rgb}\mathcal L_{\rm rgb}
 +\lambda_{\rm grad}\mathcal L_{\rm grad}
 +\lambda_{\rm color}\mathcal L_{\rm color}\\
 &+\lambda_{\rm perc}\mathcal L_{\rm perc}
 +\lambda_{\rm clip}\mathcal L_{\rm clip}.
\end{aligned}
\label{eq:ldrloss}
\end{equation}
Appendix~\ref{app:ldr-loss} gives the exact objectives. All numerical architecture choices, bounds, and loss weights are reported with the training settings in Appendix~\ref{app:implementation}.

\section{Dataset Construction and Curation}
\label{sec:data}
\begin{figure}[t]
  \centering
  \includegraphics[width=\linewidth]{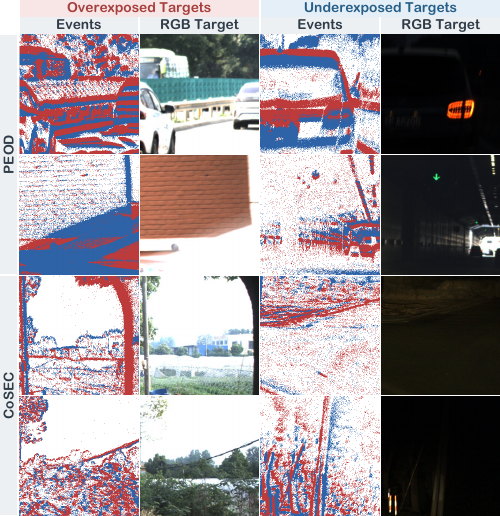}
  \caption{Excluded RGB targets with their corresponding $50$ ms event accumulations. Each pair uses the same crop and target time. RGB values are displayed without exposure adjustment.}
  \label{fig:data_curation}
\end{figure}

PEOD~\cite{Cui2026} and CoSEC~\cite{Peng2024} are event-RGB datasets developed for object detection and autonomous driving, respectively. However, neither dataset was designed specifically for supervised color reconstruction from monochrome events. Some RGB targets contain overexposed or underexposed regions that obscure surface color and texture (Figure~\ref{fig:data_curation}) and can encourage the model to learn exposure artifacts. Therefore, we construct and curate paired data from both datasets for reliable color supervision.

\noindent{\textbf{Paired Data Construction.}}
Each sample contains a $256\times256$ RGB image crop, aligned events, and an event-reconstructed grayscale image. PEOD uses a fixed temporal offset and sequence-level geometric correction, while CoSEC uses coaxially aligned RGB and event coordinates. Validity masks restrict supervision to aligned regions.

\noindent{\textbf{Data Curation.}}
For each crop sampled on a grid, we compare a short event accumulation with edges extracted from its RGB target. Crops with large residual offsets or weak edge correspondence are rejected. We also measure near-white, near-black, bright, and dark pixel fractions within each crop's valid region. Crops with excessive clipping or large bright or dark areas are excluded from training and evaluation. Retained targets keep their original RGB values. Appendix~\ref{app:curation} details sampling geometry, alignment criteria, exposure thresholds, and screening counts.

\noindent{\textbf{Retained Data.}}
The training split contains 8,068 crops from 1,755 target frames. Evaluation uses 2,220 crops from 491 target frames. All methods share the same retained evaluation targets and valid regions.

\section{Experiments and Results}
\label{sec:experiments}

\begin{table*}[t]
\centering\footnotesize
\caption{Quantitative comparison with external baselines on PEOD and CoSEC. Chr. denotes chromatic target regions, and Desat. reports the percentage of chromatic target pixels reconstructed as near-neutral. CF-MAE measures colorfulness error relative to the RGB targets.}
\label{tab:main}
\setlength{\tabcolsep}{1pt}
\begin{tabular*}{\textwidth}{@{\extracolsep{\fill}}>{\raggedright\arraybackslash}p{0.31\textwidth}|cc|cc|ccc|cc|c@{}}
\toprule
\multicolumn{1}{c|}{Method} & PSNR $\uparrow$ & SSIM $\uparrow$ & LPIPS $\downarrow$ & DISTS $\downarrow$ & Chr. $\Delta E_{00}\downarrow$ & Desat. (\%) $\downarrow$ & CF-MAE $\downarrow$ & NIQE $\downarrow$ & FID $\downarrow$ & Params. (M)\\
\midrule
\rowcolor{black!10}\multicolumn{11}{l}{\emph{Image colorization from reconstructed grayscale}}\\
Colorful Image Colorization (ECCV, 2016)~\cite{Zhang2016} & 14.306 & 0.5185 & 0.5210 & 0.3073 & 22.876 & \underline{11.114} & 19.054 & 6.766 & 207.114 & 32.24\\
DDColor (ICCV, 2023)~\cite{Kang2023} & 13.705 & 0.4919 & 0.5224 & 0.3081 & 24.317 & 11.492 & 26.032 & 6.814 & 175.876 & 227.87\\
ColorFormer (ECCV, 2022)~\cite{Ji2022ColorFormer} & 13.992 & 0.4862 & 0.4820 & 0.2975 & 21.798 & 18.978 & 18.210 & 6.850 & 138.680 & 44.82\\
BigColor (ECCV, 2022)~\cite{Kim2022BigColor} & 14.380 & 0.4887 & 0.4865 & 0.2985 & 22.365 & 17.408 & 18.944 & 6.730 & 138.394 & 585.51\\
Control Color (IJCV, 2025)~\cite{Liang2025ControlColor} & 14.214 & \underline{0.5962} & 0.4501 & \underline{0.2881} & 20.841 & 22.304 & 20.150 & 7.222 & 156.302 & 1523.61\\
L-CAD (NeurIPS, 2023)~\cite{LCAD2023} & 13.529 & 0.4858 & 0.5128 & 0.3113 & 24.229 & 16.662 & 25.378 & \underline{6.653} & 147.627 & 1103.61\\
DeOldify (GitHub, 2018)~\cite{DeOldify} & 14.483 & 0.5199 & 0.4726 & 0.3031 & 21.341 & 25.813 & 11.383 & 6.764 & 181.201 & 218.22\\
CT$^2$ (ECCV, 2022)~\cite{CT22022} & 14.314 & 0.5118 & 0.4734 & 0.3018 & 21.166 & 20.504 & 14.845 & 6.763 & 164.058 & 462.98\\
DISCO (SIGGRAPH Asia, 2022)~\cite{DISCO2022} & 14.229 & 0.5143 & 0.4783 & 0.3058 & 20.763 & 67.306 & 15.962 & 6.764 & 155.947 & 43.06\\
UniColor (SIGGRAPH Asia, 2022)~\cite{UniColor2022} & 13.601 & 0.4759 & 0.5024 & 0.3076 & 22.931 & 20.601 & 25.192 & 6.768 & 142.251 & 190.73\\
\addlinespace[2pt]
\rowcolor{black!10}\multicolumn{11}{l}{\emph{Grayscale reconstruction followed by colorization}}\\
\shortstack[l]{V2V-E2VID + DDColor\\(NeurIPS, 2025 / ICCV, 2023)~\cite{V2VE2VID2025,Kang2023}} & 14.453 & 0.4967 & 0.4567 & 0.3035 & 24.404 & \textcolor{paperred}{\textbf{10.766}} & 20.360 & 8.331 & 150.953 & 238.58\\
\addlinespace[2pt]
\rowcolor{black!10}\multicolumn{11}{l}{\emph{Event video reconstruction}}\\
EvDiff (ECCV, 2026)~\cite{EvDiff2025} & \underline{15.807} & 0.5336 & \underline{0.3856} & 0.3027 & 22.572 & 55.410 & \underline{6.837} & 9.186 & \textcolor{paperred}{\textbf{102.178}} & 2223.51\\
LongE2V (SIGGRAPH, 2026)~\cite{LongE2V2026} & 15.752 & 0.5488 & 0.4721 & 0.3150 & \underline{20.440} & 97.685 & 9.614 & 10.561 & 185.881 & 10548.57\\
\addlinespace[2pt]
\midrule
\rowcolor{paperredlight}
EventSeeColor (Ours) & \textcolor{paperred}{\textbf{16.010}} & \textcolor{paperred}{\textbf{0.6443}} & \textcolor{paperred}{\textbf{0.3329}} & \textcolor{paperred}{\textbf{0.2589}} & \textcolor{paperred}{\textbf{15.767}} & 36.603 & \textcolor{paperred}{\textbf{5.494}} & \textcolor{paperred}{\textbf{6.326}} & \underline{117.286} & 1523.98\\
\bottomrule
\end{tabular*}
\end{table*}

\subsection{Experimental Setup}
\label{sec:experimental_setup}

\noindent{\textbf{Implementation Details.}}
EventSeeColor uses aligned $256\times256$ crops and $512\times512$ inputs to the diffusion colorizer, with LoRA rank set to $4$. The auxiliary chroma loss weight is $\lambda_{\rm aux}=0.1$. BCR uses region-balancing coefficient $\beta=0.5$ and warm-region supervision weight $\lambda_w=0.2$, while LDR limits its RGB correction to $\delta=0.12$. Inference uses $20$ deterministic DDIM steps with classifier-free guidance scale $7$. Training and inference are performed on a single NVIDIA GeForce RTX 5090 GPU with 32\,GB of memory. EGF, LoRA, BCR, and LDR add 133,216, 181,248, 32,340, and 24,451 inference parameters, respectively. These additions total 371,255 parameters, increasing the Control Color backbone's parameter count by only $0.0244\%$. Appendix~\ref{app:training} provides the remaining network and optimization settings.

\noindent{\textbf{Evaluation Protocol.}}
All methods are evaluated in their native RGB output space on the curated data in Section~\ref{sec:data}. PSNR and SSIM measure image fidelity. LPIPS and DISTS~\cite{Ding2022DISTS} measure perceptual agreement, while chromatic CIEDE2000 ($\Delta E_{00}$)~\cite{Sharma2005CIEDE2000} evaluates target regions with substantial color. Desaturation (Desat.) measures the fraction of chromatic target pixels reconstructed as near-neutral. Colorfulness MAE (CF-MAE)~\cite{Hasler2003Colorfulness} measures per-crop colorfulness error, NIQE~\cite{Mittal2013NIQE} measures natural luminance statistics, and Fr\'echet Inception Distance (FID)~\cite{Heusel2017FID} compares output and target distributions. Appendix~\ref{app:metrics} defines paired metrics and aggregation, and Appendix~\ref{app:distribution} specifies global appearance and distribution metrics.

\noindent{\textbf{Comparison Baselines.}}
Table~\ref{tab:main} lists the external baselines in our comparison, organized into three categories. The first category comprises image colorization methods that receive the same normalized event-reconstructed grayscale images. For the second category, we design and construct a baseline pipeline that feeds grayscale images reconstructed by V2V-E2VID~\cite{V2VE2VID2025} into DDColor for colorization, evaluating an alternative grayscale front end. The third category comprises event video reconstruction methods with RGB frame generation capabilities, including EvDiff~\cite{EvDiff2025} and LongE2V~\cite{LongE2V2026}. We retain their temporal inference procedures and extract RGB frames corresponding to the evaluation target times. The same spatial alignment and crop rules then yield individual color-image reconstructions for comparison. Appendix~\ref{app:baselines} specifies these input and deployment settings.

\begin{figure*}[p]
  \centering
  \includegraphics[width=\textwidth]{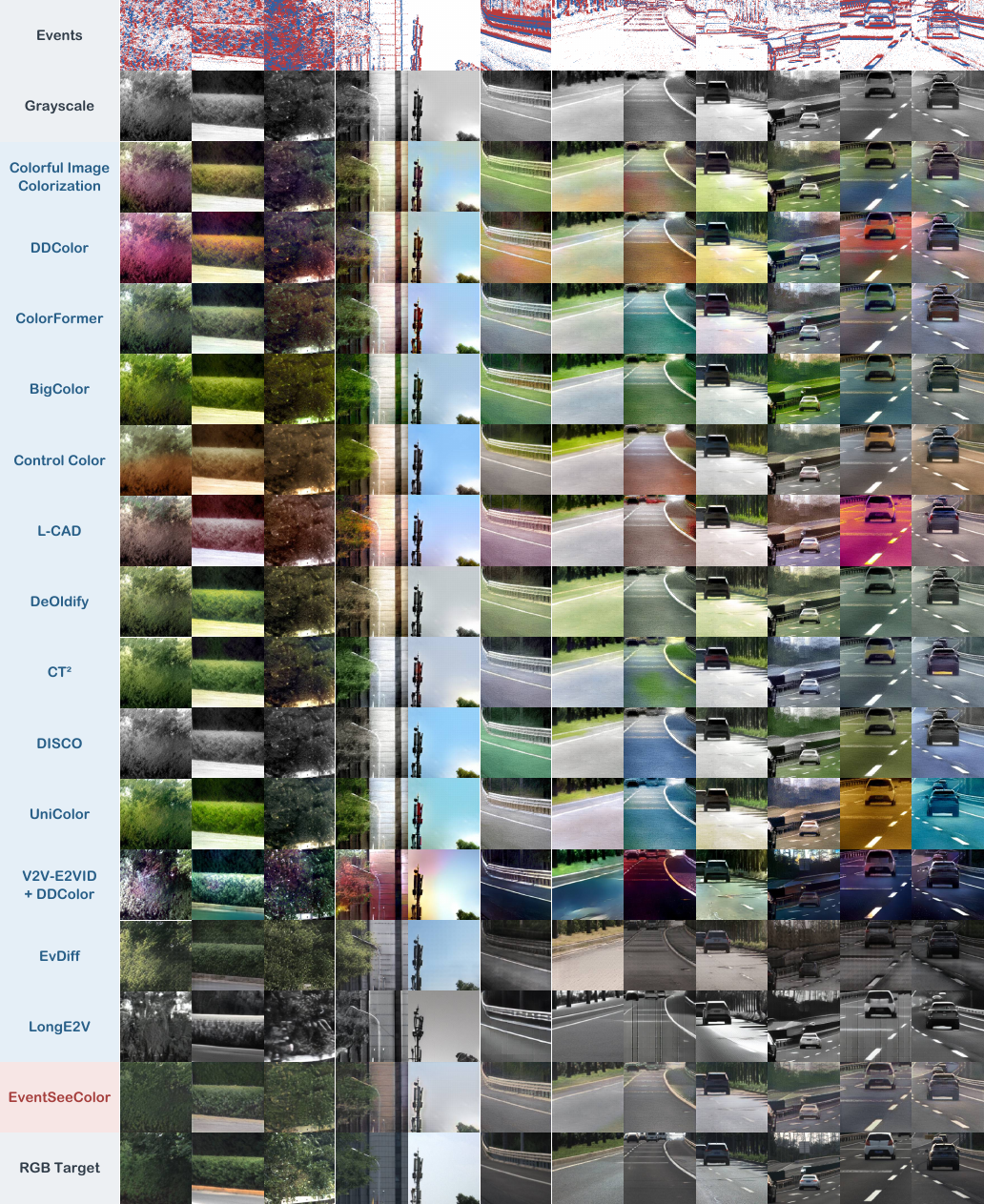}
  \caption{Qualitative comparison with the external baselines in Table~\ref{tab:main}. EventSeeColor recovers more faithful surface colors and details while reducing false coloration on neutral regions across vegetation, buildings, roads, and vehicles.}
  \label{fig:qualitative}
\end{figure*}

\begin{table*}[t]
\centering\footnotesize
\caption{Ablation study of EventSeeColor's components, event guidance, and training objectives on the evaluation set. Metrics and parameter-counting scope follow Table~\ref{tab:main}. Within each section, best and second-best quality scores are bold and underlined, respectively.}
\label{tab:ablation}
\setlength{\tabcolsep}{1pt}
\renewcommand{\arraystretch}{1.12}
\begin{tabular*}{\textwidth}{@{\extracolsep{\fill}}>{\raggedright\arraybackslash}p{0.31\textwidth}|cc|cc|ccc|cc|c@{}}
\toprule
\multicolumn{1}{c|}{Configuration} & PSNR $\uparrow$ & SSIM $\uparrow$ & LPIPS $\downarrow$ & DISTS $\downarrow$ & Chr. $\Delta E_{00}\downarrow$ & Desat. (\%) $\downarrow$ & CF-MAE $\downarrow$ & NIQE $\downarrow$ & FID $\downarrow$ & Params. (M)\\
\midrule
\rowcolor{black!10}\multicolumn{11}{l}{\emph{Components and event guidance}}\\
w/o EGF Event Path & 15.981 & 0.6415 & \underline{0.3371} & 0.2619 & \underline{15.906} & \underline{34.511} & \underline{5.597} & 6.414 & 121.923 & 1523.91\\
w/o BCR & \underline{16.042} & \underline{0.6439} & 0.3478 & 0.2638 & 16.398 & 45.884 & 6.383 & 6.416 & \textcolor{paperred}{\textbf{100.750}} & 1523.95\\
w/o LDR & 15.056 & 0.6187 & 0.4113 & 0.3104 & 16.360 & \textcolor{paperred}{\textbf{23.602}} & 6.208 & 6.424 & 151.809 & 1523.96\\
w/o Event Inputs (EGF/BCR/LDR) & \textcolor{paperred}{\textbf{16.080}} & 0.6357 & 0.3506 & \underline{0.2617} & 16.229 & 40.126 & 5.618 & \underline{6.333} & 131.985 & 1523.90\\
\midrule
\rowcolor{paperredlight}
EventSeeColor (full) & 16.010 & \textcolor{paperred}{\textbf{0.6443}} & \textcolor{paperred}{\textbf{0.3329}} & \textcolor{paperred}{\textbf{0.2589}} & \textcolor{paperred}{\textbf{15.767}} & 36.603 & \textcolor{paperred}{\textbf{5.494}} & \textcolor{paperred}{\textbf{6.326}} & \underline{117.286} & 1523.98\\
\midrule
\rowcolor{black!10}\multicolumn{11}{l}{\emph{Training objectives}}\\
w/o Warm Supervision & \underline{16.030} & 0.6355 & 0.3531 & 0.2639 & 15.987 & \underline{34.800} & 5.597 & 6.357 & 134.445 & 1523.98\\
w/o Auxiliary Chroma Supervision & 16.009 & 0.6421 & 0.3371 & 0.2608 & 15.821 & \textcolor{paperred}{\textbf{34.201}} & 5.539 & 6.377 & 120.752 & 1523.98\\
w/o LDR Perceptual Supervision & \textcolor{paperred}{\textbf{16.218}} & \textcolor{paperred}{\textbf{0.6530}} & 0.3562 & 0.2665 & \textcolor{paperred}{\textbf{15.583}} & 38.858 & 5.592 & 6.585 & 128.857 & 1523.98\\
w/o LDR Color Supervision & 16.025 & 0.6435 & 0.3360 & \textcolor{paperred}{\textbf{0.2570}} & 15.844 & 35.227 & \textcolor{paperred}{\textbf{5.486}} & \textcolor{paperred}{\textbf{6.276}} & \textcolor{paperred}{\textbf{116.688}} & 1523.98\\
w/o LDR Gradient Supervision & 16.004 & 0.6438 & \underline{0.3330} & 0.2590 & 15.772 & 36.519 & 5.499 & 6.328 & 117.445 & 1523.98\\
\midrule
\rowcolor{paperredlight}
EventSeeColor (full) & 16.010 & \underline{0.6443} & \textcolor{paperred}{\textbf{0.3329}} & \underline{0.2589} & \underline{15.767} & 36.603 & \underline{5.494} & \underline{6.326} & \underline{117.286} & 1523.98\\
\bottomrule
\end{tabular*}
\end{table*}

\subsection{Comparison with Existing Methods}
\label{sec:comparison}

\noindent{\textbf{Quantitative Comparison.}}
Table~\ref{tab:main} shows that EventSeeColor achieves the best PSNR, SSIM, LPIPS, DISTS, and chromatic CIEDE2000 among the compared methods on the combined evaluation set. In particular, EventSeeColor outperforms Control Color, which provides its pretrained diffusion backbone, in image fidelity, perceptual quality, and chromatic-region color accuracy. The lowest CF-MAE further indicates closer overall colorfulness to the targets, and the lowest NIQE indicates closer agreement with natural luminance statistics. EventSeeColor also achieves the second-lowest FID, indicating close agreement between the feature distributions of its reconstructions and real targets. Appendix~\ref{app:additional} provides additional evaluation metrics.

\begin{figure}[t]
  \centering
  \includegraphics[width=\linewidth]{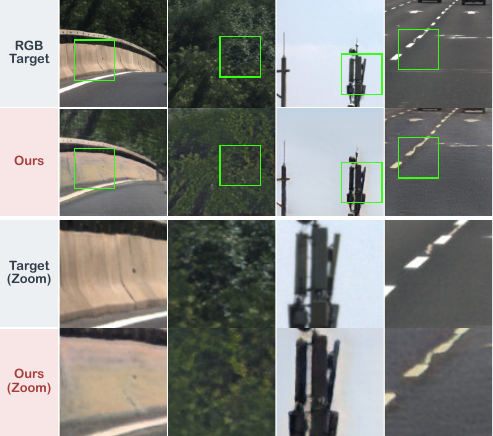}
  \caption{EventSeeColor reconstructions of concrete, foliage, sky, and road surfaces. Rows show RGB targets, outputs (Ours), and corresponding close-ups. Fluorescent green boxes mark the enlarged regions. Crops and display scales match.}
  \label{fig:color_reconstructions}
\end{figure}

\noindent{\textbf{Qualitative Comparison.}}
Figure~\ref{fig:qualitative} compares EventSeeColor with all external methods across vegetation, buildings, roads, and vehicles. Figure~\ref{fig:color_reconstructions} shows how EventSeeColor preserves distinct appearance across warm concrete, foliage, sky, and neutral road surfaces. EventSeeColor recovers foliage color while reducing false coloration on road and vehicle surfaces. Visually, EventSeeColor shows the closest color agreement with the RGB targets in the chromatic regions of these examples. In contrast, LongE2V produces nearly achromatic outputs on many colored surfaces. Appendix~\ref{app:regional_results} provides complementary regional color accuracy and desaturation metrics. Appendix~\ref{app:qualitative} provides additional qualitative comparisons of color and detail reconstruction.

\begin{figure}[t]
  \centering
  \includegraphics[width=\columnwidth]{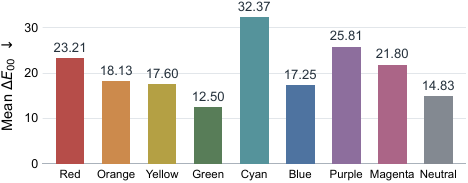}
  \caption{Target-color accuracy of EventSeeColor. Bars show crop-averaged regional CIEDE2000. Lower is better. Appendix~\ref{app:metrics} defines the groups and aggregation.}
  \label{fig:color_statistics}
\end{figure}

\noindent{\textbf{Color Recovery Analysis.}}
Figure~\ref{fig:color_statistics} reports EventSeeColor's regional CIEDE2000 by target color. Green regions have the lowest error, followed by blue and yellow among the chromatic groups.

\subsection{Ablation Studies}
\label{sec:ablation}
Table~\ref{tab:ablation} examines whether aligned events provide complementary information for colorizing event-reconstructed grayscale images. Among the component and event-guidance configurations, EventSeeColor achieves the highest SSIM and lowest LPIPS, DISTS, and chromatic error. Removing BCR increases chromatic error and desaturation, while removing LDR retains more chroma at the cost of RGB fidelity and perceptual agreement. Appendix~\ref{app:ablation_settings} defines these configurations, and Appendix~\ref{app:ablations} provides further analysis and the achromatic-input experiment. We also ablate EventSeeColor's supervision terms. Perceptual and color supervision support perceptual agreement and chromatic accuracy, respectively. Our visual assessment finds that the full configuration provides the best overall visual quality among these supervision configurations, balancing image fidelity and color accuracy (Appendix~\ref{app:qualitative_ablations}).

\section{Conclusion}
EventSeeColor reconstructs scene color from monochrome events through joint grayscale and event guidance. Region-balanced chroma supervision and local RGB refinement improve surface colors, brightness, and detail. We curate paired data from PEOD and CoSEC to support supervised color reconstruction. On the combined evaluation, EventSeeColor achieves the best structural similarity, perceptual fidelity, and chromatic-region color accuracy among the external baselines. The ablations provide empirical evidence that temporal event activity contributes to color inference beyond grayscale semantics, expanding the role of monochrome event sensing in recovering scene appearance.

\clearpage
\makeatletter
\global\@colroom=\textheight
\global\vsize=\textheight
\makeatother
\bibliographystyle{ieeenat_fullname}
\bibliography{references}
\clearpage
\appendix
\setcounter{figure}{0}
\setcounter{table}{0}
\setcounter{equation}{0}
\renewcommand{\thefigure}{A\arabic{figure}}
\renewcommand{\thetable}{A\arabic{table}}
\renewcommand{\theequation}{A.\arabic{equation}}
\renewcommand{\theHequation}{appendix.\arabic{equation}}
\begin{center}
{\large\bfseries\eventseecolortitle}\\[0.4em]
{\large Supplementary Material}
\end{center}

\section{Research Motivation}
\label{app:motivation}
Event cameras asynchronously record changes in light intensity, producing measurements of pixel location, timestamp, and polarity with high temporal resolution and a wide dynamic range~\cite{Gallego2022Survey}. Sensor models and event simulators describe how intensity trajectories, response dynamics, and threshold crossings produce event timing and polarity~\cite{Chen2026FracEvent,Chen2026EventSimulationReview,Rebecq2018ESIM,Hu2021V2E}. The resulting event stream encodes spatial structure and temporal variation for learning scene properties~\cite{Gallego2022Survey,Gehrig2019EventRepresentation}.

These measurements support 3D reconstruction~\cite{Rebecq2018EMVS,Xu2026Event3DSurvey,Xu2025Event3D}, human pose estimation~\cite{Calabrese2019DHP19,Zhou2026EventPose}, and dense optical flow estimation~\cite{Gehrig2021ERAFT}. Event-stream super-resolution improves spatial detail while preserving temporal and polarity information~\cite{Xu2026PolaritySNN}. Events also complement image measurements. Frame-event feature tracking combines image-based spatial references with asynchronous temporal updates~\cite{Gehrig2018FeatureTracking}, and event-guided frame interpolation uses motion information between captured images to reconstruct intermediate frames~\cite{Tulyakov2021TimeLens}. Under low illumination, event-guided enhancement uses structural and illumination cues from events to improve image visibility and detail~\cite{Liang2024EvLight,Zhou2026EvLIR}. These applications combine image appearance and temporal events for perception and image recovery~\cite{Liang2024EvLight,Gehrig2018FeatureTracking,Tulyakov2021TimeLens}.

Traditional event-based image reconstruction makes event recordings accessible as grayscale intensity images~\cite{Rebecq2019}. However, grayscale reconstruction does not express color differences between surfaces with similar intensity, motivating the inference of missing chromatic information~\cite{Zhang2016}. Color can distinguish the appearance of road markings, vegetation, and vehicle surfaces, and can aid recognition when objects have characteristic colors~\cite{Tanaka1999ColorDiagnosticity}. Therefore, recovering color can enrich the scene information conveyed by monochrome event recordings. Grayscale images provide spatial and semantic cues for learned colorization~\cite{Zhang2016,Kang2023}, while aligned events retain local timing, polarity, and activity~\cite{Gallego2022Survey,Gehrig2019EventRepresentation}. This complementary information motivates EventSeeColor to investigate whether temporal events improve color reconstruction beyond the spatial and semantic guidance of event-reconstructed grayscale images.

The significance of color reconstruction also extends to downstream perception. Pretrained RGB vision models can extract appearance features from reconstructed color images. Aligning and fusing these appearance features with event-derived structural and motion features could combine semantic context, surface-color distinctions, and precise temporal information within a shared representation. Such integration could improve object recognition, human pose estimation, and 3D reconstruction using monochrome event recordings as the sole sensor input. This perspective positions color reconstruction as a bridge between event sensing and appearance-aware visual understanding.

\section{Method Details}
\label{app:methoddetails}
This appendix details Event and Grayscale Fusion (EGF), Balanced Chroma Refinement (BCR), Local Detail Refinement (LDR), and their inputs and training objectives. Symbols follow Section~\ref{sec:method}: $G_t$ is the normalized event-reconstructed grayscale image, $V_t$ the aligned event tensor, $A_t$ the event activity map, $m_t$ the validity mask, and $L_t$ the calibrated event lightness. The fused hint $H_t$ and fixed photographic prompt $\pi$ guide the diffusion colorizer. The Lab chroma vector $\chroma=(a,b)$ has magnitude $C=\|\chroma\|_2$. BCR refines initial chroma $\chroma_t^0$ into $\chroma_t^r$ and renders the intermediate RGB image $I_t^r$. LDR produces the unclipped RGB output $\widetilde I_t$ and final RGB image $\widehat I_t$. The time subscript $t$ is omitted where a single target time is understood. A superscript $*$ denotes a training or evaluation target.

\subsection{Event Representation and Lightness Calibration}
\label{app:inputs}

\noindent{\textbf{Reconstruction and Event Encoding.}}
The official E2VID lightweight recurrent checkpoint reconstructs a grayscale image at each target time from the preceding one second of events through twenty consecutive 50\,ms updates, each encoded with five signed temporal bins. Recurrent state is reset before each target history and retained across its twenty updates. The target timestamp is expressed in the aligned event clock. The event-reconstructed grayscale image $G_t$ is normalized using its first and ninety-ninth percentiles. The aligned event tensor $V_t$ covers the final 50\,ms ending at that target timestamp and uses ten temporal bins for each polarity. Each event contributes to its adjacent bins through linear interpolation. Two polarity count maps and two last-timestamp surfaces are appended, producing a 24-channel tensor. For a nonnegative voxel or count value $x$, normalization is:
\begin{equation}
  \mathcal N_\tau(x)=\min\!\left(\frac{\log(1+x)}{\log(1+\tau)},1\right).
\end{equation}
The voxel and count scales are given in Table~\ref{tab:hyperparameters}. Each timestamp surface records the most recent event of its polarity within the interval, normalized to $[0,1]$. The activity map $A_t$ is the mean of the two normalized count channels. Adaptive max pooling of the nonzero support $A_t>0$ gives the binary EGF support map $S_\ell$ at each feature scale. LDR receives the continuous activity map $A_t$.

\noindent{\textbf{Lightness Calibration.}}
The calibrated event lightness $L_t$ is derived from the E2VID grayscale reconstruction before percentile normalization. Replicating the raw grayscale into RGB and converting it to Lab gives $L_t^{\rm raw}$. A dataset-specific affine map is fitted by least squares with a nonnegative slope over valid training pixels:
\begin{equation}
  L_t=\operatorname{clip}(\mu_d L_t^{\rm raw}+\nu_d,0,100).
  \label{eq:calibration}
\end{equation}
Here, $d$ indexes the dataset. The fitted coefficients are $(\mu_d,\nu_d)=(0.991864,-0.467356)$ for PEOD and $(0.546846,12.793549)$ for CoSEC. These fitted coefficients remain fixed during training and inference. Lightness calibration determines the RGB targets used to train the diffusion colorizer and the lightness used to render BCR outputs. LDR subsequently adjusts the rendered RGB image, including its lightness.

\noindent{\textbf{Gamut Mapping.}}
For lightness $L$ and predicted chroma $\chroma$, $\Pi_L$ finds the largest feasible scale $\xi\in[0,1]$ such that $\mathcal T(L,\xi\chroma)$ lies in the RGB gamut. The implementation uses bisection along the chroma direction. This operation preserves Lab lightness and hue while reducing only out-of-gamut chroma. Gamut mapping produces the initial chroma $\chroma^0$ supplied to BCR and maps the refined chroma $\chroma^r$ before rendering the intermediate RGB image $I^r$. The final RGB correction follows Eq.~\ref{eq:output}.

\subsection{Fusion and Fixed Prompt Settings}
\label{app:prompt}

\noindent{\textbf{Event and Grayscale Fusion.}}
EGF uses three encoder stages with separate grayscale and event branches. For $256\times256$ inputs, the stages operate at $256\times256$, $128\times128$, and $64\times64$, and the fused hint has $320$ channels at $64\times64$ resolution. Grayscale is resized to the event resolution by area interpolation, and both inputs are multiplied by the validity mask before encoding. Each stage consists of a convolution, GroupNorm, and SiLU. The gate in Eq.~\ref{eq:gate} uses a pointwise convolution on concatenated features. Features from all scales are pooled to the resolution of the pretrained grayscale hint $H_t^0$ before projection. The grayscale and event projections are separate pointwise convolutions, and the event projection omits its bias. Zero initialization of both projections makes the fused hint $H_t$ equal to $H_t^0$ at the start of adaptation.

\noindent{\textbf{Fixed Photographic Prompt.}}
Training and inference use the same photographic prompt ``natural color photograph, realistic materials, balanced saturation, neutral lighting,'' followed by the quality suffix ``best quality, detailed, real.'' The resulting text condition $\pi$ is shared across all images. Sampling uses the fixed negative prompt ``a black and white photo, longbody, lowres, bad anatomy, bad hands, missing fingers, extra digit, fewer digits, cropped, worst quality, low quality.''

\subsection{Balanced Chroma Refinement}
\label{app:refinement}
BCR receives the five-channel image input, with $s$ the chroma normalization scale:
\begin{equation}
  X=[G,\chroma^0/s,\|\chroma^0\|_2/s,m].
\end{equation}
The image encoder processes $X\odot m$ through two stride-two stages, producing $64\times64$ features for a $256\times256$ crop. The event tensor $V\odot m$ is resized by area interpolation to the image feature resolution and processed by its own encoder. The image and event features are concatenated and processed by a joint context network to produce features $F$. Two heads produce a two-channel gain precursor $q$ and residual $\mathbf r$:
\begin{equation}
\begin{aligned}
  q&=\tanh(\mathcal U(U_{\rm gain}(F))),\\
  \boldsymbol\alpha&=\mathbf1+\gamma_+\max(q,0)+\gamma_-\min(q,0),\\
  \mathbf r&=\rho\tanh(\mathcal U(U_{\rm res}(F))),
\end{aligned}
\end{equation}
where $U_{\rm gain}$ and $U_{\rm res}$ are the two prediction heads and $\mathcal U$ bilinearly upsamples their logits to the output resolution before the nonlinearities. The gain $\boldsymbol\alpha$ and chroma residual $\mathbf r$ are used in Eq.~\ref{eq:bcr}. The scalars $\gamma_+$ and $\gamma_-$ set the amplification and attenuation ranges, and $\rho$ sets the residual bound. Independent gain channels permit different multiplicative adjustments of $a$ and $b$. Both heads have zero weights and biases at initialization, giving $\boldsymbol\alpha=\mathbf1$ and $\mathbf r=\mathbf0$.

\subsection{Local Detail Refinement}
\label{app:ldr}
LDR receives the six-channel tensor $[I^r,G,A,m]\odot m$. Two full-resolution convolutions produce shallow features. The context branch takes these shallow features as input and applies a stride-two convolution followed by a dilated convolution. Bilinear upsampling returns the context features to the shallow resolution, and their concatenation is decoded into three RGB residual logits. The final convolution has zero weights and biases at initialization. Applying the bounded residual in Eq.~\ref{eq:output} preserves the intermediate RGB image $I^r$ outside the valid region. The correction acts in encoded sRGB coordinates, so all three output channels can respond to local appearance errors. Table~\ref{tab:hyperparameters} gives the widths and residual bound.

\subsection{Training Objectives}
\label{app:objectives}
Throughout the objectives, $j$ indexes images, $i$ indexes pixels, and $k\in\{a,b\}$ indexes chroma channels.

\noindent{\textbf{Diffusion Denoising.}}
The training target is RGB rendered from paired target chroma at calibrated event lightness after gamut mapping. At diffusion step $n$, the target latent is perturbed by Gaussian noise $\epsilon$ to give $z_n$. Let $\widehat\epsilon=\epsilon_\theta(z_n,n,H_t,\pi)$ be the noise predicted by the denoiser with parameters $\theta$, $m^z$ the validity mask resized by area interpolation to latent resolution, and $d_z$ the latent channel count. The denoising loss is:
\begin{equation}
  \mathcal L_{\rm diff}=\mathbb E\!\left[
  \frac{\sum_i m_i^z\|\epsilon_i-\widehat\epsilon_i\|_2^2}
       {d_z\sum_i m_i^z}\right].
\end{equation}

\noindent{\textbf{Auxiliary Chroma Supervision.}}
The training-only auxiliary chroma head predicts normalized coarse target chroma $\mathbf u=h_{\rm aux}(H_t)$. With adaptive average pooling $\mathcal P$, the valid area and target chroma at hint resolution are:
\begin{equation}
  \bar m=\mathcal P(m),\qquad
  \bar\chroma^*=\frac{\mathcal P(m\odot\Pi_L(\chroma^*))}
                        {\max(\bar m,\varepsilon)},
\end{equation}
where $\varepsilon=10^{-8}$ stabilizes division at empty locations. Define $\omega_i=\bar m_i[1+\eta\min(\|\bar\chroma_i^*\|_2/\kappa,1)]$. The auxiliary objective is:
\begin{equation}
  \mathcal L_{\rm aux}=\frac{\sum_i\omega_i\sum_k
      \operatorname{SmoothL1}(u_{ik}-\bar c^*_{ik}/s)}
      {\max(2\sum_i\omega_i,\varepsilon)}.
\end{equation}
Thus, valid area determines target pooling and loss weighting.

\noindent{\textbf{Region-Balanced Chroma Objective.}}
\label{app:bcrloss}
BCR is trained against the original paired target Lab chroma. The diffusion stage uses gamut-mapped RGB targets for latent encoding, while BCR directly supervises chroma and applies gamut projection when rendering RGB. The per-pixel regression loss and chroma weight are:
\begin{equation}
\begin{aligned}
  \ell_{ji}&=\frac12\sum_k\operatorname{SmoothL1}
        \!\left(\frac{c^r_{jik}-c^*_{jik}}{s}\right),\\
  w_{ji}&=1+\eta\min(C^*_{ji}/\kappa,1).
\end{aligned}
\label{eq:pixelloss}
\end{equation}
SmoothL1 uses unit transition width. The scale $s$ normalizes Lab chroma, while $\eta$ and $\kappa$ control additional weight for chromatic targets.

\noindent{\textbf{Global and Region Terms.}}
Target chroma thresholds $\tau_n$ and $\tau_c$ define the neutral, transitional, and chromatic subsets by $C^*<\tau_n$, $\tau_n\le C^*\le\tau_c$, and $C^*>\tau_c$. The reductions follow Eq.~\ref{eq:regionloss}. The global term divides weighted error by the valid pixel count. The region term first averages within each target subset and then gives equal weight to nonempty subsets and to images.

\noindent{\textbf{Chroma Shortfall and Neutral-Chroma Supervision.}}
Writing $\Omega_C$ and $\Omega_N$ for all chromatic and neutral pixels in the batch, respectively, the shortfall and neutral terms are:
\begin{equation}
  \mathcal L_{\rm short}=\frac1{|\Omega_C|}\sum_{(j,i)\in\Omega_C}
  \frac{(C^*_{ji}-C^r_{ji})_+}{s},
\end{equation}
\begin{equation}
  \mathcal L_{\rm neutral}=\frac1{|\Omega_N|}\sum_{(j,i)\in\Omega_N}
  \frac{\|\chroma^r_{ji}-\chroma^*_{ji}\|_2^2}{2s^2},
\end{equation}
where $(x)_+=\max(x,0)$ and $C^r=\|\chroma^r\|_2$. The neutral term matches the target's small chroma, retaining weak surface coloration.

\noindent{\textbf{Warm-Region Supervision.}}
Target hue in degrees is $h^*=(180/\pi)\operatorname{atan2}(b^*,a^*)$. The warm subset $\Omega_{jW}$ satisfies $C^*>\tau_c$ and $h^*\in[h_-,h_+]$. For the set $\mathcal J_W$ of images containing this subset, warm-region supervision is:
\begin{equation}
  \mathcal L_{\rm warm}=\frac1{|\mathcal J_W|}
  \sum_{j\in\mathcal J_W}\frac1{|\Omega_{jW}|}
  \sum_{i\in\Omega_{jW}}\ell_{ji}.
\end{equation}
This hue-specific mean gives warm surface colors direct supervision within the chromatic subset, complementing the magnitude-based region balancing. Warm-region supervision uses unweighted regression errors. Empty region terms contribute zero. The complete BCR objective follows Eq.~\ref{eq:bcrloss}, with values given in Table~\ref{tab:training-settings}.

\subsection{Local Detail Refinement Objectives}
\label{app:ldr-loss}
The LDR objective supervises the clipped RGB output $\widehat I$ and penalizes clipping of the unclipped RGB output $\widetilde I$. Let $I^*$ denote the paired RGB target and define $\phi(x)=\sqrt{x^2+\varepsilon_{\rm rgb}^2}$. For any multichannel image $Z$, $\langle Z\rangle_m$ denotes the average over its valid spatial entries and all channels, aggregated across the batch. The reconstruction and gradient terms are:
\begin{equation}
\begin{aligned}
  \mathcal L_{\rm rgb}&=\langle\phi(\widehat I-I^*)\rangle_m,\\
  \mathcal L_{\rm grad}&=\frac12\sum_{d\in\{x,y\}}
  \langle\phi(D_d\widehat I-D_d I^*)\rangle_{m^d},
\end{aligned}
\label{eq:ldr-rgb}
\end{equation}
where $D_d$ is a forward finite difference and $m^d$ is the product of the validity masks at the two adjacent pixels. The gradient term supervises changes across neighboring pixels, complementing pointwise RGB fidelity.

LDR color supervision $\mathcal L_{\rm color}$ uses the Lab chroma of the final RGB prediction $\widehat I$. For image $j$ and pixel $i$, define $\ell^{\rm RGB}_{ji}=\frac12\sum_k\operatorname{SmoothL1}((\widehat c_{jik}-c^*_{jik})/s)$, where $\widehat\chroma$ is obtained from $\widehat I$. The target chroma defines the same neutral, transitional, and chromatic subsets $\Omega_{jr}$ used by BCR. Each nonempty subset contributes equally:
\begin{equation}
  \mathcal L_{\rm color}=\frac1B\sum_j\frac1{|\mathcal R_j|}
  \sum_{r\in\mathcal R_j}\frac1{|\Omega_{jr}|}
  \sum_{i\in\Omega_{jr}}\ell^{\rm RGB}_{ji}.
\label{eq:ldr-color}
\end{equation}
This objective uses the unweighted chroma regression error within each region. Before differentiable Lab conversion, RGB values are lower-bounded by $10^{-6}$ to keep the gradient finite at black pixels.

LDR perceptual supervision $\mathcal L_{\rm perc}$ uses LPIPS with a frozen AlexNet network~\cite{Zhang2018LPIPS}. For each image, the final RGB prediction $\widehat I$ and paired RGB target $I^*$ are cropped to the valid bounding rectangle and rescaled from $[0,1]$ to $[-1,1]$. The perceptual term is the mean LPIPS distance over images. The clipping penalty $\mathcal L_{\rm clip}$ measures the squared RGB difference introduced by clipping output values to $[0,1]$:
\begin{equation}
  \mathcal L_{\rm clip}=\langle(\widetilde I-\widehat I)^2\rangle_m.
\end{equation}
These terms form Eq.~\ref{eq:ldrloss}. Their weights and the Charbonnier constant are reported in Table~\ref{tab:training-settings}.

\subsection{Optional Grayscale Scene Prompt}
\label{app:optional_gsp}
EventSeeColor uses the fixed photographic prompt $\pi$ in Appendix~\ref{app:prompt} by default. An optional Grayscale Scene Prompt (GSP) augments this prompt with image-specific semantic context. A frozen Qwen2.5-VL-3B-Instruct model~\cite{Bai2025} describes visible objects, spatial arrangement, and recognizable materials from the event-reconstructed grayscale image. Explicit color words are removed before the description is concatenated with the fixed photographic prompt. GSP uses no RGB reference. EGF, BCR, LDR, and the sampling settings are unchanged.

\noindent{\textbf{Color Retention and Reconstruction Accuracy.}}
The configurations with and without GSP use the same training seed and budgets, with BCR trained on the corresponding diffusion colorizer outputs and LDR trained on the corresponding BCR outputs. On the 2,220 evaluation samples, adding GSP reduces desaturation from $36.603\%$ to $34.531\%$ and changes PSNR from $16.010$ to $16.023$ dB. However, LPIPS increases from $0.3329$ to $0.3375$, and chromatic-region chroma RMSE increases from $8.894$ to $9.031$. Therefore, EventSeeColor uses the fixed photographic prompt.

\noindent{\textbf{Additional Computation.}}
GSP adds $3{,}754.62$ million frozen parameters and no additional trainable parameters. The diffusion sampler retains 20 DDIM steps. On an NVIDIA GeForce RTX 5090 GPU, generating descriptions for 2,220 crops took $537.36$ seconds with batch size four, with an amortized captioning-stage cost of $0.242$ seconds per crop, including model loading, preprocessing, generation, and cache writing. Cached descriptions can be reused for repeated colorization of the same inputs.

\section{Datasets and Metrics}
\label{app:data}
\subsection{Paired Data and Sequence Splits}
\label{app:splits}
PEOD~\cite{Cui2026} and CoSEC~\cite{Peng2024} supply real event recordings with corresponding RGB images. PEOD provides the main collection of aligned scene observations, while CoSEC adds a distinct acquisition setting with traffic scenes. Table~\ref{tab:split} gives the retained sequence, frame, and crop counts. The split is disjoint by recording sequence within each dataset. PEOD evaluation uses sequences 036, 063, 070, 078, and 096. CoSEC evaluation uses City\_Day\_000.

The retained evaluation set consists of crops from the same fixed target frames for every method. Each method's output is mapped to those crop coordinates, and scores are measured on the original rectangular valid region. These rectangles cover between $96.51\%$ and $100\%$ of a crop.

\begin{table}[t]
\centering\footnotesize
\caption{Data used for training and evaluation. Sequences are disjoint between the two partitions within each dataset.}
\label{tab:split}
\setlength{\tabcolsep}{2pt}
\begin{tabular*}{\columnwidth}{@{\extracolsep{\fill}}l|c|ccc@{}}
\toprule
\multicolumn{1}{c|}{Dataset} & Partition & Sequences & Frames & Crops\\
\midrule
PEOD & Training & 18 & 1,685 & 7,927\\
     & Evaluation & 5 & 479 & 2,198\\
CoSEC & Training & 3 & 70 & 141\\
      & Evaluation & 1 & 12 & 22\\
\midrule
Total & Training & 21 & 1,755 & 8,068\\
      & Evaluation & 6 & 491 & 2,220\\
\bottomrule
\end{tabular*}
\end{table}

\subsection{Native RGB Evaluation}
\label{app:native}
Evaluation measures framewise reconstruction accuracy from each method's native RGB output, including its predicted lightness. Image colorizers receive the same normalized event-reconstructed grayscale crop, except for the V2V-E2VID and DDColor pipeline. Direct event reconstruction methods use their released procedures to process events. Their event histories and temporal context are specified in Appendix~\ref{app:baselines}. All methods are compared against the same aligned RGB targets, with no output fitting to evaluation images. Method categories in the tables distinguish their input and inference routes.

Each crop receives equal weight in the overall mean. Therefore, combined scores reflect the dataset composition in Table~\ref{tab:split}. A regional score contributes only when the corresponding target region is present. There are 2,076 crops with chromatic support, 1,905 with warm support, and 2,220 with neutral support.

\subsection{Reference Image and Regional Color Metrics}
\label{app:metrics}

\noindent{\textbf{Image Fidelity.}}
PSNR is reported in dB and uses mean squared error in sRGB with a unit signal range. SSIM~\cite{Wang2004SSIM} uses an $11\times11$ Gaussian window with standard deviation $1.5$, population covariance, and exclusion of the window border. LPIPS uses the calibrated AlexNet model with RGB scaled to $[-1,1]$~\cite{Zhang2018LPIPS}. DISTS measures learned structural and textural agreement~\cite{Ding2022DISTS}. Both perceptual metrics operate on the valid rectangle without metric-side resizing. LPIPS is included in Local Detail Refinement training, while DISTS supplies an additional perceptual assessment.

\noindent{\textbf{Chroma and Perceptual Color Difference.}}
Evaluation converts predicted and target sRGB to CIELAB under D65. For a target-defined region $\Omega$, the chroma root mean squared error is:
\begin{equation}
\operatorname{RMSE}_{ab}(\Omega)=
\sqrt{\frac{1}{2|\Omega|}\sum_{i\in\Omega}
\|\chroma_i^{\rm pred}-\chroma_i^*\|_2^2}.
\label{eq:metric-ab}
\end{equation}
This metric isolates the two chroma channels. CIEDE2000 additionally accounts for lightness and perceptual nonuniformity~\cite{Sharma2005CIEDE2000}. Its score is the mean pixelwise $\Delta E_{00}$ with $k_L=k_C=k_H=1$, computed over all valid pixels or the specified target region. Chromatic pixels satisfy $C^*>10$, and neutral pixels satisfy $C^*<5$, where $C^*=\|\chroma^*\|_2$. Warm pixels are chromatic pixels with target hue between $-30^\circ$ and $75^\circ$.

\noindent{\textbf{Target Color Groups.}}
Figure~\ref{fig:color_statistics} in the main paper uses eight chromatic groups defined by the RGB target's HSV hue. The intervals are red $[345^\circ,360^\circ)\cup[0^\circ,15^\circ)$, orange $[15^\circ,45^\circ)$, yellow $[45^\circ,75^\circ)$, green $[75^\circ,165^\circ)$, cyan $[165^\circ,195^\circ)$, blue $[195^\circ,255^\circ)$, purple $[255^\circ,285^\circ)$, and magenta $[285^\circ,345^\circ)$. Each interval includes its lower boundary and excludes its upper boundary. Chromatic groups require Lab $C^*>10$, and a separate neutral group uses $C^*<5$. The transition range $5\leq C^*\leq10$ is excluded from this grouping. A crop supports a group when that region occupies at least $1\%$ of its valid pixels. Pixelwise CIEDE2000 is first averaged within the region, then averaged equally over supporting crops.

\noindent{\textbf{Desaturation and False Color.}}
Desaturation (Desat.) measures missed color as the fraction of chromatic target pixels whose predicted chroma magnitude is below $5$. Neutral false color measures the fraction of neutral target pixels with predicted chroma magnitude above $20$. Warm leakage further restricts those predictions to the warm hue interval. Each fraction is computed within a crop and then averaged over supported crops. Together, the measurements distinguish suppressed target color from color introduced on neutral surfaces.

\subsection{Distribution and Naturalness Metrics}
\label{app:distribution}

\noindent{\textbf{FID and KID.}}
Fr\'echet Inception Distance (FID)~\cite{Heusel2017FID} and Kernel Inception Distance (KID)~\cite{Binkowski2018KID} compare the reconstructed-image distribution with the paired RGB target distribution. Predictions and targets are cropped to identical valid rectangles and quantized to 8-bit RGB. Features are extracted with the TensorFlow-compatible Inception-V3 weights and internal $299\times299$ preprocessing in \texttt{torch-fidelity}. FID uses all 2,220 outputs and targets. KID uses 100 independent subset pairs of size 1,000 and a cubic polynomial kernel, with $\gamma=1/2048$ and unit offset. Tables report the KID subset mean and standard deviation multiplied by $10^3$. This standard deviation describes variation across the sampled subsets.

\noindent{\textbf{Colorfulness.}}
Colorfulness summarizes opponent-channel variation~\cite{Hasler2003Colorfulness}. Following the implementation used in DDColor, the evaluation computes absolute opponent channels $r_g=|R-G|$ and $y_b=|(R+G)/2-B|$ from 8-bit RGB. The score is:
\begin{equation}
\mathrm{CF}=\sqrt{\sigma_{r_g}^2+\sigma_{y_b}^2}
+0.3\sqrt{\mu_{r_g}^2+\mu_{y_b}^2},
\label{eq:colorfulness}
\end{equation}
where $\mu$ and $\sigma$ are the pixelwise mean and standard deviation. Colorfulness mean absolute error (CF-MAE) averages the absolute predicted-to-target CF difference per crop.

\noindent{\textbf{Natural Image Statistics.}}
The Natural Image Quality Evaluator (NIQE)~\cite{Mittal2013NIQE} and Blind/Referenceless Image Spatial Quality Evaluator (BRISQUE)~\cite{Mittal2012BRISQUE} use the pretrained defaults in \texttt{pyiqa}. NIQE operates on YIQ luminance and uses complete $96\times96$ blocks, discarding incomplete blocks along the bottom and right boundaries. BRISQUE uses its standard luminance input. NIQE and BRISQUE measure agreement with natural-image statistics independently of the paired RGB target. Paired perceptual and regional color metrics measure reconstruction accuracy against that target. Reference RGB images are scored with the same settings to characterize natural-image statistics in the evaluated data.

\subsection{Data Construction and Curation Details}
\label{app:curation}
\noindent{\textbf{Candidate Sampling.}}
PEOD combines fixed sample locations and additional spatial grid locations within the same sequence partition. The fixed-location pool uses four deterministically selected $256\times256$ crops per sampled frame. Additional grid crops have top-left coordinates $x\in\{0,256,512,768,1024\}$ and $y\in\{0,232,464\}$. A grid crop is omitted if its intersection-over-union with an exposure-qualified fixed-location crop exceeds $0.5$. PEOD uses a fixed $20$ ms RGB-to-event timestamp offset and the sequence-level spatial correction. The associated validity mask excludes warped borders with a two-pixel erosion.

CoSEC samples every fifth RGB frame, starting from timestamps with at least one second of available event history. Eight nonoverlapping $256\times256$ crops use top-left coordinates $x\in\{88,344,600,856\}$ and $y\in\{56,312\}$. RGB and events use the supplied coaxial coordinate system. Sequence assignments are fixed before crop selection. Table~\ref{tab:split} summarizes the retained sequence, frame, and crop counts.

\noindent{\textbf{RGB Exposure Screening.}}
Statistics are computed only on valid pixels of the original RGB target. Let $r,g,b\in[0,255]$ denote its encoded sRGB channel values, with encoded luma $Y_{\rm sRGB}=0.2126r+0.7152g+0.0722b$. A crop is rejected if any of the following conditions holds:
\begin{itemize}
  \item At least $10\%$ of valid pixels are near-white, with $\min(r,g,b)\geq250$.
  \item At least $10\%$ are near-black, with $\max(r,g,b)\leq10$.
  \item At least $20\%$ have low encoded luma, with $Y_{\rm sRGB}\leq32$.
  \item At least $20\%$ are bright across all channels, with $\min(r,g,b)\geq235$.
\end{itemize}
The same rules apply to training and evaluation. Rejection removes the entire crop from the paired set. Retained targets keep their recorded RGB values, and selection does not use model predictions or reconstruction scores.

\noindent{\textbf{Crop-Level Correspondence.}}
Grid candidates also undergo local alignment screening. The event edge map marks pixels with at least one event in the final $5$ ms before the target. For RGB scaled to $[0,1]$, grayscale conversion and a $3\times3$ Sobel operator provide gradient magnitudes. The RGB edge threshold is the $85$th percentile of magnitudes greater than $0.02$. Each edge map must contain at least $64$ supporting pixels. A search within $\pm6$ pixels identifies the best translation. The crop is retained only if its residual displacement is at most $1.5$ pixels, its zero-shift edge cosine similarity is at least $0.15$, and that similarity exceeds the strongest $16$-pixel horizontal or vertical shift control by at least $0.02$. The search tests correspondence after the fixed calibration. It does not introduce a per-crop warp into the training targets.

\noindent{\textbf{Screening Counts.}}
Exposure screening retains 2,260 of the 9,976 fixed-location PEOD crops. Among 78,072 additional PEOD grid candidates, 20,406 pass the exposure rules and 7,865 also pass the local correspondence checks. Among 1,488 CoSEC grid candidates, 746 pass exposure screening and 163 pass both stages. Across these candidate pools, exposure screening excludes 66,124 crops, and alignment screening excludes a further 13,124 exposure-qualified crops. The remaining 10,288 crops comprise the 8,068 training and 2,220 evaluation samples in Table~\ref{tab:split}.

\section{Experimental Settings}
\label{app:implementation}
\subsection{EventSeeColor}
\label{app:training}
Diffusion colorizer adaptation updates EGF, the auxiliary chroma head, and LoRA parameters in ControlNet text cross-attention~\cite{Hu2022LoRA}. The pretrained diffusion colorizer weights remain fixed. BCR is then trained with the adapted diffusion colorizer frozen. Finally, LDR is trained from the frozen BCR RGB outputs, paired grayscale, event activity, and validity masks. Only the loss functions access the paired color targets during LDR training. All three stages use the training partition in Table~\ref{tab:split}. Training and inference use an NVIDIA RTX 5090 GPU. The exact network and optimization choices are given in Tables~\ref{tab:hyperparameters} and~\ref{tab:training-settings}.

At inference, the auxiliary chroma head is omitted and sampling uses DDIM~\cite{Song2021DDIM}. Normalized grayscale is enlarged to $512\times512$ using Lanczos interpolation. The frozen content-guided deformable VAE from Control Color~\cite{Liang2025ControlColor} decodes the sampled latent using grayscale content features. The decoded RGB image is resized to the original crop resolution using Lanczos interpolation before chroma extraction and gamut mapping for BCR. Grayscale provides the initial intensity structure, and event features guide both color generation and refinement. The final output is the native RGB prediction from LDR. Appendix~\ref{app:efficiency} reports parameter counts and LDR latency.

Self-attention guidance follows Control Color~\cite{Liang2025ControlColor} and operates during early reverse diffusion down to timestep 600 on the original 1,000-step timeline. The 20 DDIM steps count sampling updates along this timeline.

\begin{table*}[t]
\centering\footnotesize
\caption{Network and inference settings. Lightness calibration is specified in Appendix~\ref{app:inputs}.}
\label{tab:hyperparameters}
\setlength{\tabcolsep}{2pt}
\begin{tabular*}{\textwidth}{@{\extracolsep{\fill}}>{\raggedright\arraybackslash}p{0.17\textwidth}|>{\raggedright\arraybackslash}p{0.37\textwidth}|>{\centering\arraybackslash}p{0.43\textwidth}@{}}
\toprule
\multicolumn{1}{c|}{Component} & \multicolumn{1}{c|}{Setting} & \multicolumn{1}{c}{Value}\\
\midrule
Inputs & Aligned crop / diffusion image & $256\times256$ / $512\times512$\\
       & Event bins per polarity / total channels & $10$ / $24$\\
       & Voxel scale / count scale & $\tau_{\rm voxel}=16$, $\tau_{\rm count}=64$\\
\midrule
EGF & Stage widths / strides & $(16,32,64)$ / $(1,2,2)$\\
    & Stage operator / normalization & $3\times3$ convolution, 4-group GroupNorm, SiLU\\
    & Gate and projection kernels / hint size & $1\times1$ / $320\times64\times64$\\
\midrule
LoRA & Attention projections / rank / alpha & Query, key, value, output / $4$ / $4$\\
\midrule
Auxiliary chroma head & Pointwise channel widths & $320\to32\to2$, intermediate SiLU\\
\midrule
Text condition & Prompt & Fixed across images, Appendix~\ref{app:prompt}\\
    & CLIP input length & $77$ tokens including padding\\
\midrule
BCR & Image branch widths / strides & $5\to16\to32$ / $(2,2)$\\
    & Event branch / context widths & $24\to16$ / $48\to32\to32$\\
    & Context dilation / output heads & $(1,2)$ / two $32\to2$ pointwise convolutions\\
    & Gain parameters / residual bound & $\gamma_+=0.5$, $\gamma_-=1$, $\rho=12$\\
\midrule
LDR & Shallow widths / context widths & $6\to16\to16$ / $16\to32\to32$\\
    & Context strides / dilation & $(2,1)$ / $(1,2)$\\
    & Residual decoder / operator & $48\to16\to3$ / $3\times3$ convolution\\
    & Activation / RGB correction bound & SiLU / $\delta=0.12$\\
\midrule
Sampling & DDIM steps / stochasticity & $20$ / $0$\\
         & Classifier-free guidance / hint scale & $7$ / $1$\\
         & Self-attention guidance / timestep cutoff & $0.05$ / $600$\\
\midrule
Rendering & Chroma gamut bisection iterations & $18$\\
\bottomrule
\end{tabular*}
\end{table*}

\begin{table*}[t]
\centering\footnotesize
\caption{Training settings and objective weights. Region thresholds are expressed in Lab chroma units.}
\label{tab:training-settings}
\setlength{\tabcolsep}{2pt}
\begin{tabular*}{\textwidth}{@{\extracolsep{\fill}}>{\raggedright\arraybackslash}p{0.17\textwidth}|>{\raggedright\arraybackslash}p{0.37\textwidth}|>{\centering\arraybackslash}p{0.43\textwidth}@{}}
\toprule
\multicolumn{1}{c|}{Component} & \multicolumn{1}{c|}{Setting} & \multicolumn{1}{c}{Value}\\
\midrule
Chroma objectives & Chroma scale / pixel weighting & $s=32$, $\eta=2$, $\kappa=20$\\
          & Region thresholds / warm hue & $\tau_n=5$, $\tau_c=10$, $[h_-,h_+]=[-30^\circ,75^\circ]$\\
          & Auxiliary chroma / region weights & $\lambda_{\rm aux}=0.1$, $\beta=0.5$\\
          & Shortfall / neutral / warm weights & $\lambda_s=0.2$, $\lambda_n=0.5$, $\lambda_w=0.2$\\
\midrule
LDR objective & RGB / gradient / color weights & $\lambda_{\rm rgb}=1$, $\lambda_{\rm grad}=0.1$, $\lambda_{\rm color}=0.25$\\
          & Perceptual / clipping weights & $\lambda_{\rm perc}=0.1$, $\lambda_{\rm clip}=0.1$\\
          & Charbonnier constant & $\varepsilon_{\rm rgb}=10^{-3}$\\
\midrule
Diffusion colorizer & Epochs / batch / accumulation & $4$ / $1$ / $4$\\
          & AdamW learning rates & LoRA $10^{-5}$, EGF and auxiliary chroma head $10^{-4}$\\
          & Weight decay & $0$\\
\midrule
BCR & Epochs / batch / optimizer & $20$ / $16$ / AdamW\\
    & Learning rate / weight decay & $10^{-4}$ / $10^{-4}$\\
\midrule
LDR & Epochs / batch / optimizer & $20$ / $8$ / AdamW\\
    & Learning rate / weight decay & $10^{-4}$ / $10^{-4}$\\
\midrule
All stages & Gradient clipping / training seed & $1$ / $3407$\\
\bottomrule
\end{tabular*}
\end{table*}

\subsection{Comparison Baselines}
\label{app:baselines}
All external models use their released pretrained weights. Each model produces one prediction per target. The temporal context and RGB frame extraction for the video reconstruction baselines are specified below. The image colorizers use the same E2VID reconstruction, normalized between its first and ninety-ninth percentiles, and follow their released input conversion and RGB postprocessing. We restore their predictions to the common crop coordinates and score native RGB within the valid region. RGB targets are used only for evaluation. Stochastic image colorizers use seed 3407 separately for each crop.

\noindent{\textbf{Parameter Counts.}}
\label{app:parameter_counts}
Table~\ref{tab:main} in the main paper reports all parameters in the networks used for inference, including frozen weights and excluding training-only heads. Image colorizers and EventSeeColor exclude the shared E2VID front end. Image colorizer counts also exclude independent caption generation. V2V-E2VID + DDColor includes both reconstruction and colorization networks, and EvDiff includes its event encoder. LongE2V includes the T5 text encoder even though its empty-prompt embedding is cached, with fused adaptation weights counted only once.

\noindent{\textbf{Colorful Image Colorization.}}
The released model~\cite{Zhang2016} operates at $256\times256$ and combines predicted chroma with the input image's Lab lightness.

\noindent{\textbf{DDColor.}}
The large ModelScope checkpoint~\cite{Kang2023} operates at $512\times512$ and combines predicted chroma with the input image's Lab lightness.

\noindent{\textbf{ColorFormer.}}
The released generator~\cite{Ji2022ColorFormer} uses the GLH backbone and color and semantic memories at $256\times256$.

\noindent{\textbf{BigColor.}}
The epoch-11 exponential-moving-average generator~\cite{Kim2022BigColor} operates at $256\times256$. Its grayscale classifier selects the highest-probability class, and a Gaussian latent with standard deviation 0.8 supplies the generative input. The released grayscale-lightness fusion is retained.

\noindent{\textbf{Control Color.}}
The original diffusion model and deformable decoder~\cite{Liang2025ControlColor} operate at $512\times512$, with 20 DDIM steps, guidance scale 7, and deterministic DDIM sampling. Control Color and L-CAD share scene descriptions generated from grayscale by a frozen vision-language model~\cite{Bai2025}. These descriptions provide neither target colors nor RGB references.

\noindent{\textbf{L-CAD.}}
The COCO language checkpoint~\cite{LCAD2023} operates at $256\times256$, with 50 DDIM steps, guidance scale 4.5, and DDIM noise parameter zero. Its released Lab lightness normalization and standard sampling procedure are retained.

\noindent{\textbf{DeOldify.}}
The stable ResNet-101 generator and official colorization filter~\cite{DeOldify} use render factor 35, corresponding to a $560\times560$ model input, with postprocessing enabled.

\noindent{\textbf{CT$^2$.}}
The released ViT-large checkpoint~\cite{CT22022} operates at $256\times256$, with 313 color tokens, four luminance-prior bins, and classification decoding at temperature 0.38. Its auxiliary chroma inputs are set to zero.

\noindent{\textbf{DISCO.}}
The recommended beta checkpoint~\cite{DISCO2022} operates at $256\times256$, with eight anchors, 16 superpixels, hidden width 64, dense positions, and the enhanced output head. Its test-time sampling temperature is zero. Its auxiliary chroma inputs are set to zero.

\noindent{\textbf{UniColor.}}
The ImageNet checkpoint~\cite{UniColor2022} operates at $256\times256$ in automatic mode, with empty strokes and no reference image. Sampling uses a $16\times16$ token grid, 256 autoregressive steps, and top-$100$ selection. The official input conversion and color resizing are retained.

\noindent{\textbf{V2V-E2VID + DDColor.}}
V2V-E2VID~\cite{V2VE2VID2025} uses the released epoch-77 reconstruction checkpoint trained on 10,000 videos, with unnormalized five-bin event voxels and continuous recurrent history. Each reconstructed target crop is percentile-normalized and colorized by the frozen DDColor model specified above. Native frame intervals are processed continuously, including non-target frames.

\noindent{\textbf{EvDiff.}}
Native event streams pass through the recurrent five-bin event encoder~\cite{EvDiff2025}, followed by the released SD3 Medium model with rank-64 adaptation, bundled prompt embeddings, and a one-step flow update at timestep 680. Evaluation uses FP32 arithmetic and the released stochastic sampling behavior with seed 3407. The recurrent state, including the temporal-fusion state, is reset at sequence boundaries and carried through every intervening frame before the target. EvDiff and V2V-E2VID + DDColor use the $20\,000\,\mu\mathrm{s}$ PEOD timestamp offset and native CoSEC timestamps.

\noindent{\textbf{LongE2V.}}
The released adapter and CogVideoX-5B-I2V backbone~\cite{LongE2V2026} operate in reconstruction mode. Three-bin signed event voxels undergo the released nonzero normalization and are resized to $480\times720$. Sampling uses the empty text prompt, 20 denoising steps, guidance scale 0, BF16 arithmetic, and seed 6666 at each sequence boundary. Each block reconstructs 49 frames from 48 event intervals, with 20 context frames and adaptive context switching at threshold 0.05 with one retry. All intervals are processed from the start of each sequence. The initial reference input is a constant image whose latents are discarded by the reconstruction path. Subsequent context uses generated grayscale frames. Evaluation uses RGB outputs before grayscale conversion, with grayscale feedback preserved. For target frames inside each block, reconstruction also uses later events from that block. Postprocessing that fits brightness to target-image statistics is omitted. Outputs are resized bilinearly to the native sensor geometry before applying the common alignment and crop. The resulting native RGB outputs are evaluated under the protocol in Appendix~\ref{app:native}.

\subsection{Recent Baselines Not Included}
\label{app:baseline_availability}
SeAda~\cite{Du2025SeAda}, Beyond Fixed Luminance~\cite{Maheshwari2026BeyondLuminance}, Color-Turbo~\cite{Li2025ColorTurbo}, and ColorFLUX~\cite{Li2026ColorFLUX} are excluded from the quantitative comparison because publicly accessible official pretrained checkpoints could not be located as of 15 October 2026.

\subsection{Ablation Settings}
\label{app:ablation_settings}
The component and event-guidance rows in Table~\ref{tab:ablation} in the main paper compare complete trained configurations. In \emph{w/o EGF Event Path}, the EGF event branch is removed while BCR and LDR retain their event inputs. In \emph{w/o BCR}, the entire BCR stage, including chroma refinement and rendering at calibrated event lightness, is bypassed. The diffusion colorizer's RGB output is passed directly to a retrained LDR. In \emph{w/o Event Inputs (EGF/BCR/LDR)}, the event branches in EGF and BCR and the event activity channel in LDR are removed. The remaining grayscale input is still reconstructed from events. The grayscale-only diffusion colorizer and BCR use the same training budgets as their counterparts with event inputs, followed by training the corresponding LDR. In \emph{w/o LDR}, evaluation uses the intermediate BCR RGB image $I^r$. All component configurations share data, validity masks, spatial alignment, and RGB targets.

The training-objective rows use the supervision names in Section~\ref{sec:objectives}. \emph{w/o Warm Supervision} sets the BCR warm-region weight $\lambda_w$ to zero. \emph{w/o Auxiliary Chroma Supervision} removes the training-only auxiliary chroma head and its loss $\mathcal L_{\rm aux}$. The three LDR supervision configurations set $\lambda_{\rm perc}$, $\lambda_{\rm color}$, or $\lambda_{\rm grad}$ to zero, respectively. These weights correspond to LDR perceptual, color, and gradient supervision in Eq.~\ref{eq:ldrloss}.

The achromatic-input experiment replaces the BCR RGB image $I^r$ with an achromatic image rendered at the same calibrated event lightness $L$. LDR retains $G$, $A$, and $m$ as inputs and can learn bounded RGB corrections. The LDR training schedule and objective match those of the other LDR configurations.

\subsection{Efficiency Measurement}
\label{app:efficiency}

\noindent{\textbf{Parameter Overhead.}}
Under the counting scope in Appendix~\ref{app:parameter_counts}, Control Color contains 1,523,608,805 parameters and EventSeeColor contains 1,523,980,060. The difference is 371,255 parameters, or $0.0244\%$ of Control Color's inference parameters. These additions comprise 133,216 parameters in EGF, 181,248 in LoRA, 32,340 in BCR, and 24,451 in LDR. The auxiliary chroma head adds 10,338 parameters during training and is omitted at inference. All pretrained diffusion weights remain frozen during adaptation.

\noindent{\textbf{Sampling and Timing.}}
EventSeeColor and the Control Color baseline both use $512\times512$ diffusion inputs, 20 DDIM steps, and classifier-free guidance scale 7. BCR and LDR operate at the $256\times256$ crop resolution. LDR timing uses FP32 arithmetic and batch size one on an NVIDIA RTX 5090, with GPU synchronization immediately before and after each forward pass. Excluding the first invocation, its mean latency over 2,219 evaluation crops is $0.886$\,ms. Inputs are already on the GPU when timing starts. This measurement covers LDR alone and excludes grayscale reconstruction, diffusion sampling, BCR, data transfer, file I/O, and metric computation.

\section{Additional Results}
\label{app:additional}

\subsection{Distribution and Appearance}
Table~\ref{tab:distribution} complements the paired reconstruction metrics with distribution statistics and full-image color differences. EvDiff leads FID and KID, followed by EventSeeColor. EventSeeColor obtains the lowest CF-MAE, showing closer per-image colorfulness to the corresponding target. EvDiff and LongE2V obtain the lowest full-image CIEDE2000 scores. Combined with Table~\ref{tab:regions}, this result highlights the contribution of lightness and large neutral regions to a whole-image color difference. EventSeeColor also achieves the lowest NIQE, while several image colorizers have lower BRISQUE scores.

\subsection{Regional Color Accuracy}
\label{app:regional_results}
Table~\ref{tab:regions} reports chroma RMSE separately for chromatic, neutral, and warm target regions. EventSeeColor has the lowest chromatic-region and warm-region chroma RMSE while limiting strong warm coloration of neutral pixels. Methods with lower desaturation can introduce larger chroma errors or stronger false color, so desaturation is interpreted together with the corresponding chroma RMSE columns. The nearly achromatic LongE2V achieves the lowest full-image and neutral-region chroma RMSE while discarding much of the color in target chromatic regions.

\begin{figure*}[!t]
\centering
\includegraphics[width=\textwidth]{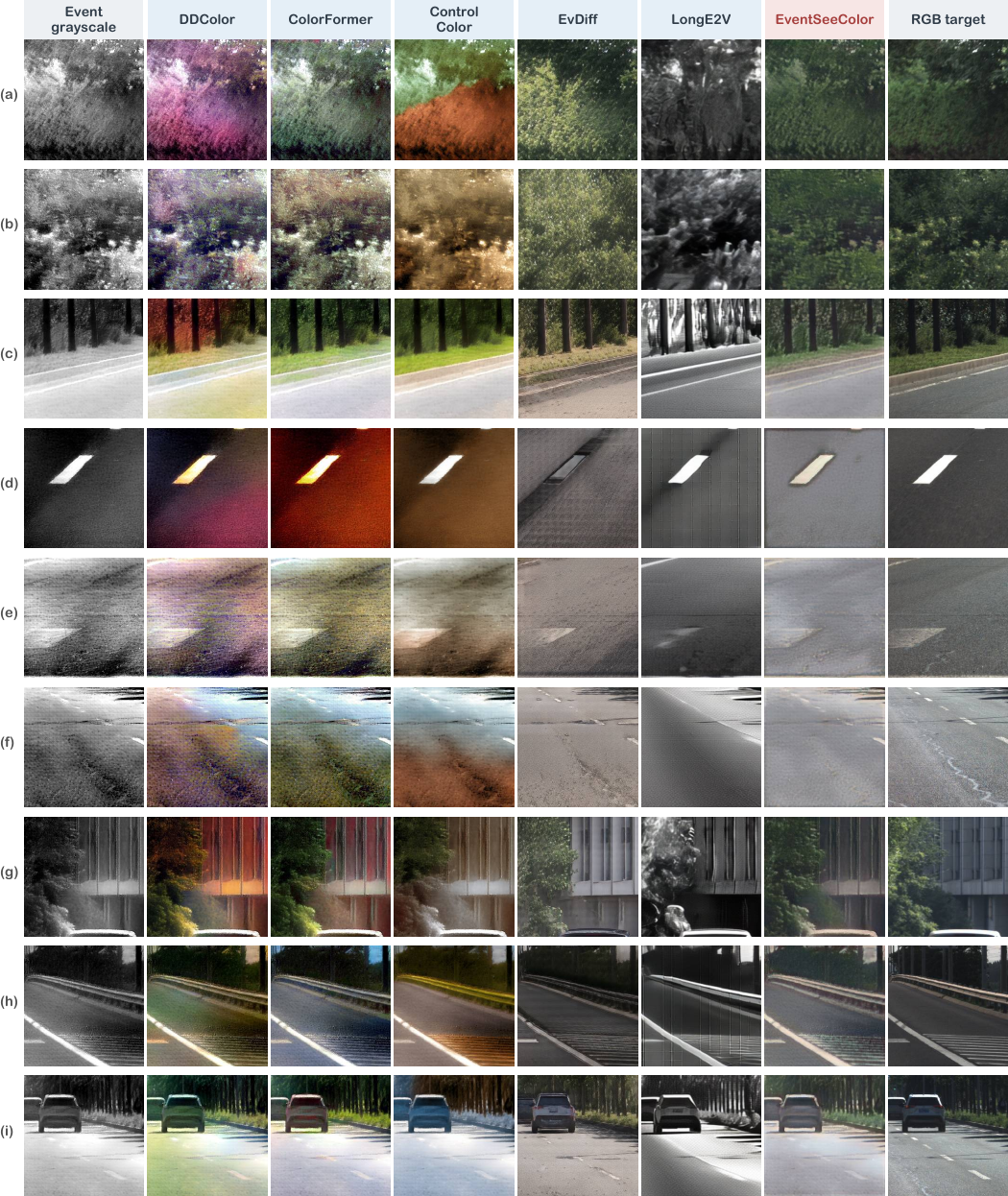}
\caption{Additional comparisons with external baselines. Rows (a) - (c) show vegetation, (d) - (f) road markings and texture, and (g) - (i) a facade, guardrail, and vehicle. Native RGB outputs use identical crop coordinates and display scales, with the RGB target as reference.}
\label{fig:qualitative_extended}
\end{figure*}

\begin{table*}[t]
\centering\footnotesize
\caption{Distribution, full-image color difference, and auxiliary appearance metrics on native RGB outputs. KID mean and subset standard deviation are multiplied by $10^3$. Lower values are better for all metrics. The same input groups as Table~\ref{tab:main} in the main paper apply.}
\label{tab:distribution}
\setlength{\tabcolsep}{1pt}
\begin{tabular*}{\textwidth}{@{}>{\raggedright\arraybackslash}p{0.25\textwidth}|>{\centering\arraybackslash}p{0.08\textwidth}>{\centering\arraybackslash}p{0.14\textwidth}|>{\centering\arraybackslash}p{0.10\textwidth}>{\centering\arraybackslash}p{0.10\textwidth}>{\centering\arraybackslash}p{0.12\textwidth}|>{\centering\arraybackslash}p{0.08\textwidth}>{\centering\arraybackslash}p{\dimexpr0.13\textwidth-14\tabcolsep-3\arrayrulewidth\relax}@{}}
\toprule
\multicolumn{1}{c|}{Method} & FID $\downarrow$ & KID $\times10^3$ $\downarrow$ & CF-MAE $\downarrow$ & All $\Delta E_{00}$ $\downarrow$ & Neutral $\Delta E_{00}$ $\downarrow$ & NIQE $\downarrow$ & BRISQUE $\downarrow$\\
\midrule
\rowcolor{black!10}\multicolumn{8}{l}{\emph{Image colorization from reconstructed grayscale}}\\
Colorful Image Colorization & 207.114 & 151.511 $\pm$ 2.959 & 19.054 & 22.490 & 21.671 & 6.766 & 14.711\\
DDColor & 175.876 & 110.633 $\pm$ 2.765 & 26.032 & 23.284 & 22.345 & 6.814 & 14.688\\
ColorFormer & 138.680 & 78.946 $\pm$ 2.359 & 18.210 & 20.664 & 20.126 & 6.850 & \underline{14.378}\\
BigColor & 138.394 & 80.011 $\pm$ 2.439 & 18.944 & 21.000 & 20.380 & 6.730 & 14.469\\
Control Color & 156.302 & 94.606 $\pm$ 2.868 & 20.150 & 20.301 & 19.762 & 7.222 & 27.166\\
L-CAD & 147.627 & 86.338 $\pm$ 2.482 & 25.378 & 22.493 & 21.719 & \underline{6.653} & \textcolor{paperred}{\textbf{14.156}}\\
DeOldify & 181.201 & 116.005 $\pm$ 2.593 & 11.383 & 20.358 & 19.813 & 6.764 & 14.400\\
CT$^2$ & 164.058 & 109.108 $\pm$ 2.767 & 14.845 & 19.958 & 19.548 & 6.763 & 14.706\\
DISCO & 155.947 & 95.145 $\pm$ 2.325 & 15.962 & 18.964 & 18.117 & 6.764 & 14.422\\
UniColor & 142.251 & 85.176 $\pm$ 2.596 & 25.192 & 21.949 & 21.353 & 6.768 & 14.948\\
\rowcolor{black!10}\multicolumn{8}{l}{\emph{Grayscale reconstruction followed by colorization}}\\
V2V-E2VID + DDColor & 150.953 & 93.694 $\pm$ 2.789 & 20.360 & 20.996 & 20.106 & 8.331 & 30.597\\
\rowcolor{black!10}\multicolumn{8}{l}{\emph{Event video reconstruction}}\\
EvDiff & \textcolor{paperred}{\textbf{102.178}} & \textcolor{paperred}{\textbf{53.138}} $\pm$ 1.920 & \underline{6.837} & \textcolor{paperred}{\textbf{11.846}} & \textcolor{paperred}{\textbf{10.969}} & 9.186 & 28.092\\
LongE2V & 185.881 & 116.852 $\pm$ 3.394 & 9.614 & \underline{13.618} & \underline{12.025} & 10.561 & 40.801\\
\midrule
\rowcolor{paperredlight}
EventSeeColor (Ours) & \underline{117.286} & \underline{61.307} $\pm$ 2.422 & \textcolor{paperred}{\textbf{5.494}} & 15.101 & 14.828 & \textcolor{paperred}{\textbf{6.326}} & 28.134\\
\bottomrule
\end{tabular*}
\end{table*}

\begin{table*}[t]
\centering\footnotesize
\caption{Chroma error and regional color behavior of native RGB outputs. Chroma RMSE occupies the first four numeric columns. The final columns report desaturation, neutral false color, and warm color leakage in percent. Lower is better.}
\label{tab:regions}
\setlength{\tabcolsep}{1pt}
\begin{tabular*}{\textwidth}{@{}>{\raggedright\arraybackslash}p{0.25\textwidth}|>{\centering\arraybackslash}p{0.07\textwidth}>{\centering\arraybackslash}p{0.08\textwidth}>{\centering\arraybackslash}p{0.095\textwidth}>{\centering\arraybackslash}p{0.08\textwidth}|>{\centering\arraybackslash}p{0.105\textwidth}>{\centering\arraybackslash}p{0.15\textwidth}>{\centering\arraybackslash}p{\dimexpr0.17\textwidth-14\tabcolsep-2\arrayrulewidth\relax}@{}}
\toprule
\multicolumn{1}{c|}{Method} & All $ab$ $\downarrow$ & Chr. $ab$ $\downarrow$ & Neutral $ab$ $\downarrow$ & Warm $ab$ $\downarrow$ & Desat. (\%) $\downarrow$ & False color (\%) $\downarrow$ & Warm leak (\%) $\downarrow$\\
\midrule
\rowcolor{black!10}\multicolumn{8}{l}{\emph{Image colorization from reconstructed grayscale}}\\
Colorful Image Colorization & 14.839 & 14.252 & 13.963 & 12.632 & \underline{11.114} & 34.237 & 6.347\\
DDColor & 16.170 & 16.626 & 15.267 & 14.954 & 11.492 & 38.603 & 21.012\\
ColorFormer & 11.870 & 13.402 & 11.221 & 14.960 & 18.978 & 18.761 & 8.791\\
BigColor & 14.921 & 15.956 & 14.050 & 16.398 & 17.408 & 33.901 & 1.575\\
Control Color & 12.662 & 13.687 & 11.824 & 12.962 & 22.304 & 27.229 & 17.823\\
L-CAD & 15.026 & 17.727 & 14.057 & 16.814 & 16.662 & 33.114 & 23.802\\
DeOldify & 10.893 & 12.376 & 10.131 & 12.835 & 25.813 & 19.430 & 1.313\\
CT$^2$ & 11.699 & 13.738 & 10.927 & 14.699 & 20.504 & 22.523 & 2.217\\
DISCO & 10.326 & 12.225 & 8.872 & 13.515 & 67.306 & 20.039 & 4.499\\
UniColor & 15.487 & 16.110 & 14.723 & 17.239 & 20.601 & 36.801 & 8.508\\
\rowcolor{black!10}\multicolumn{8}{l}{\emph{Grayscale reconstruction followed by colorization}}\\
V2V-E2VID + DDColor & 13.967 & 16.600 & 12.864 & 16.239 & \textcolor{paperred}{\textbf{10.766}} & 27.901 & 10.627\\
\rowcolor{black!10}\multicolumn{8}{l}{\emph{Event video reconstruction}}\\
EvDiff & 4.791 & \underline{9.411} & 3.639 & \underline{11.423} & 55.410 & 2.035 & 0.016\\
LongE2V & \textcolor{paperred}{\textbf{4.671}} & 11.049 & \textcolor{paperred}{\textbf{2.175}} & 12.432 & 97.685 & \textcolor{paperred}{\textbf{0.012}} & \underline{0.007}\\
\midrule
\rowcolor{paperredlight}
EventSeeColor (Ours) & \underline{4.777} & \textcolor{paperred}{\textbf{8.894}} & \underline{3.602} & \textcolor{paperred}{\textbf{10.907}} & 36.603 & \underline{0.057} & \textcolor{paperred}{\textbf{0.002}}\\
\bottomrule
\end{tabular*}
\end{table*}

\begin{figure*}[!t]
\centering
\includegraphics[width=0.92\textwidth]{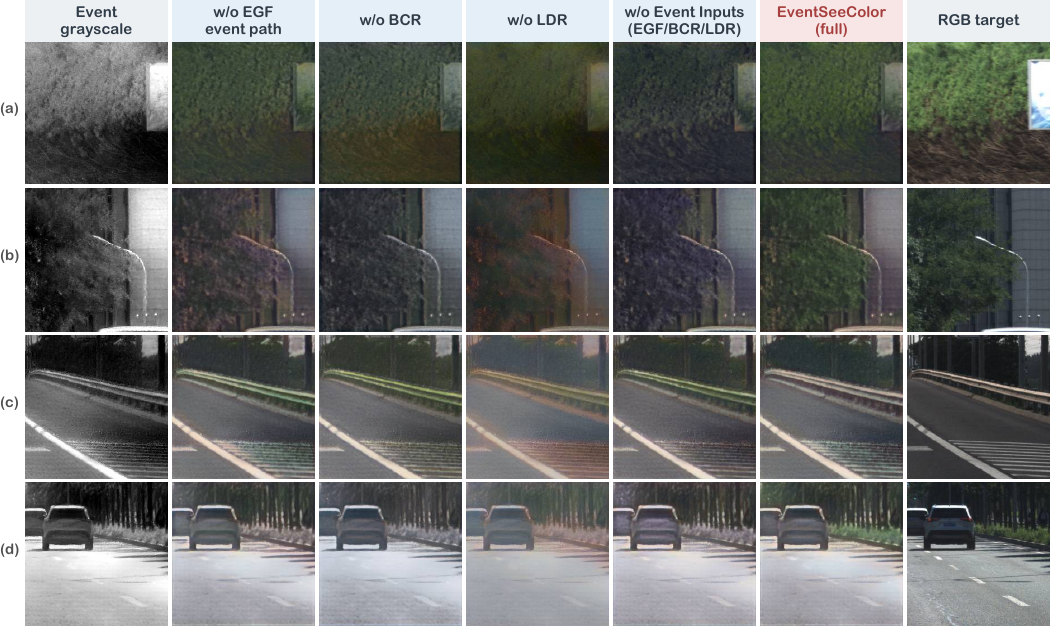}
\caption{Qualitative component and event-guidance comparisons. The configurations correspond to Table~\ref{tab:ablation} in the main paper. Rows (a) and (b) show foliage, row (c) shows a guardrail, and row (d) shows a roadside vehicle. Each row uses the same crop and scale.}
\label{fig:qualitative_components}

\vspace{10pt}
\includegraphics[width=0.92\textwidth]{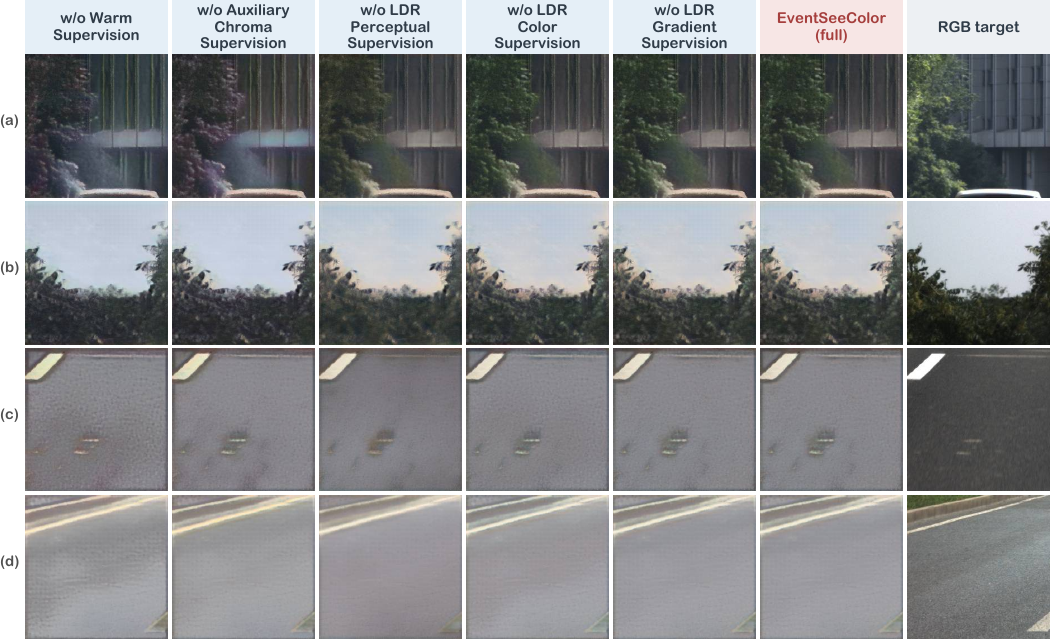}
\caption{Qualitative supervision comparisons. Rows (a) and (b) show foliage, building, and sky regions. Rows (c) and (d) show road markings and texture. Each row uses the same crop and scale, with the full EventSeeColor configuration and RGB target in the final two columns.}
\label{fig:qualitative_supervision}
\end{figure*}

\subsection{Component, Event Guidance, and Supervision Results}
\label{app:ablations}
\noindent{\textbf{Data Adaptation and Fusion.}}
Table~\ref{tab:data_adaptation} separates paired-data adaptation from additional fusion and includes the complete method as a reference. The three adapted Control Color variants start from the same weights and use 8,068 training crops, four epochs, 8,068 optimizer updates, and seed 3407. LoRA updates 181,248 parameters. The grayscale and event--grayscale fusion controls each add a 97,376-parameter adapter and evaluate the diffusion output before BCR or LDR. These four Control Color conditions share grayscale-derived scene descriptions and diffusion settings. Their predicted chroma is rendered with the same calibrated event lightness and gamut mapping. The final row applies this common rendering and scoring procedure to the complete EventSeeColor output. Its native RGB results, including the lightness predicted by LDR, are reported in Table~\ref{tab:main} in the main paper.

LoRA alone reduces $ab$ RMSE by $24.21\%$ relative to frozen Control Color. Adding a grayscale adapter reduces the error by a further $3.45\%$, and event--grayscale fusion improves upon the grayscale adapter by $3.22\%$. Relative to LoRA alone, event--grayscale fusion reduces $ab$ RMSE by $6.56\%$ and also improves PSNR and SSIM. These results show that fusion provides reconstruction gains beyond paired-data adaptation alone. The complete EventSeeColor configuration further improves $ab$ RMSE, approximate $\Delta E_{76}$, and PSNR under this common scoring protocol, while the event--grayscale fusion control retains the highest SSIM.

\begin{table}[t]
\centering\footnotesize
\caption{Data adaptation, fusion, and the complete method with common fixed-lightness outputs on 2,220 evaluation crops. Best and second-best values are bold and underlined.}
\label{tab:data_adaptation}
\setlength{\tabcolsep}{2pt}
\begin{tabular*}{\columnwidth}{@{}>{\raggedright\arraybackslash}p{0.39\columnwidth}|*{4}{>{\centering\arraybackslash}p{\dimexpr0.1525\columnwidth-2\tabcolsep-0.25\arrayrulewidth\relax}}@{}}
\toprule
\multicolumn{1}{c|}{Configuration} & \shortstack{$ab$ RMSE\\$\downarrow$} & \shortstack{Approx.\\$\Delta E_{76}\downarrow$} & \shortstack{PSNR\\$\uparrow$} & \shortstack{SSIM\\$\uparrow$}\\
\midrule
Frozen Control Color & 12.8797 & 26.0718 & 14.3973 & 0.6007\\
LoRA only & 9.7619 & 23.5487 & 14.6786 & 0.6193\\
LoRA + gray adapter & 9.4248 & 23.1856 & 14.7649 & \underline{0.6194}\\
LoRA + event--gray fusion & \underline{9.1218} & \underline{22.9109} & \underline{14.8167} & \textcolor{paperred}{\textbf{0.6209}}\\
\midrule
\rowcolor{paperredlight}
EventSeeColor (full) & \textcolor{paperred}{\textbf{4.7769}} & \textcolor{paperred}{\textbf{19.5217}} & \textcolor{paperred}{\textbf{15.1163}} & 0.6176\\
\bottomrule
\end{tabular*}
\end{table}

\noindent{\textbf{Full-Pipeline Configurations.}}
EventSeeColor has lower chromatic CIEDE2000, higher SSIM, and lower LPIPS and DISTS than the configuration without event inputs to EGF, BCR, and LDR in Table~\ref{tab:ablation} in the main paper. That configuration has higher PSNR. Removing only the EGF event path produces smaller differences from EventSeeColor and a lower desaturation rate.

The achromatic-input experiment achieves PSNR 16.086, SSIM 0.6497, and LPIPS 0.3313, but its chromatic-region chroma RMSE is 10.368 and its desaturation rate is 77.574\%. EventSeeColor achieves chromatic-region chroma RMSE 8.894 and desaturation 36.603\%. These measurements distinguish whole-image similarity from successful target-color recovery. The configuration without LDR has the lowest desaturation in Table~\ref{tab:ablation} in the main paper, while EventSeeColor has stronger RGB fidelity and perceptual agreement.

The supervision comparisons in the same table clarify the roles of the LDR objectives. Perceptual supervision improves LPIPS and DISTS, while color supervision reduces chromatic CIEDE2000. Removing perceptual supervision improves PSNR and SSIM, and removing color supervision improves DISTS and FID, reflecting the different optimization targets. These complementary gains, together with the visual comparisons in Appendix~\ref{app:qualitative_ablations}, support their joint use to balance regional color accuracy and local detail.

\subsection{Additional Qualitative Comparisons}
\label{app:qualitative}
Figure~\ref{fig:qualitative_extended} compares native RGB outputs on foliage and road scenes. In rows (a) - (c), EventSeeColor recovers green vegetation while preserving the separation between foliage, tree trunks, and neutral pavement. DDColor, ColorFormer, and Control Color introduce different magenta, cyan, or warm casts. EvDiff produces green foliage with fine texture, although the reconstructed contours and local texture can differ from the target. LongE2V remains largely achromatic in these examples.

Rows (d) - (f) examine road markings and asphalt. EventSeeColor retains the marking geometry and a relatively neutral road appearance. Several colorization baselines apply strong colors to these surfaces. EvDiff can produce smooth road texture but changes some marking intensities and shapes, while LongE2V introduces regular texture patterns. Compared with the RGB targets, fine road texture in the EventSeeColor outputs is smoother, with differences in local brightness.

Rows (g) - (i) extend the comparison to a building facade, a guardrail, and a vehicle. EventSeeColor separates green vegetation from relatively neutral structures, while several colorizers introduce strong casts on the facade, rail, or vehicle body.

Appendix~\ref{app:regional_results} quantifies regional color accuracy and desaturation, complementing these visual comparisons.

\subsection{Qualitative Component and Supervision Comparisons}
\label{app:qualitative_ablations}
Figure~\ref{fig:qualitative_components} visualizes the component and event-guidance configurations in Table~\ref{tab:ablation} in the main paper. In rows (a) and (b), EventSeeColor retains green foliage, while several configurations produce weaker green or a purple or brown cast. The configuration without BCR produces a pronounced color shift in these examples. Without LDR, the images are softer and local contrast is lower. In row (c), EventSeeColor limits green tint on the guardrail while preserving its boundary. Row (d) shows the separation between roadside vegetation, the vehicle, and the road.

Figure~\ref{fig:qualitative_supervision} compares the supervision configurations. Rows (a) and (b) show foliage and sky color differences when BCR warm-region supervision or auxiliary chroma supervision is removed. Removing LDR perceptual supervision softens leaf boundaries, building detail, and the small road patches in rows (c) and (d). Removing LDR color supervision changes color intensity, whereas removing LDR gradient supervision produces subtler visual differences in these examples. Our visual assessment of these supervision configurations finds that the full EventSeeColor configuration provides the best overall visual quality through its combined color fidelity and local detail.

\end{document}